\documentclass[journal]{IEEEtran}
\IEEEoverridecommandlockouts
\usepackage{cite}
\usepackage{amsmath,amssymb,amsfonts}
\usepackage{algorithmic}
\usepackage{graphicx}
\usepackage{textcomp}
\usepackage{xcolor}
\usepackage{subfigure}
\usepackage{orcidlink}
\usepackage{mars}

\PassOptionsToPackage{hyphens}{url}\usepackage{hyperref}

\title{
ShanghaiTech Mapping Robot is All You Need: Robot System for Collecting Universal Ground Vehicle Datasets\\
}

\author{Bowen Xu~\orcidlink{0009-0004-4325-4876},
	Xiting Zhao~\orcidlink{0009-0003-1138-1726},
	Delin Feng~\orcidlink{0009-0005-2092-0159},
	Yuanyuan Yang~\orcidlink{0000-0001-9624-8585},
	and S\"oren Schwertfeger~\orcidlink{0000-0003-2879-1636}~\IEEEmembership{Senior Member,~IEEE}
\thanks{This work was supported by the Science and Technology Commission of Shanghai Municipality (STCSM), project 22JC1410700 "Evaluation of real-time localization and mapping algorithms for intelligent robots". This work has also been partially funded by the Shanghai Frontiers Science Center of Human-centered Artificial Intelligence. The experiments of this work were supported by the core facility Platform of Computer Science and Communication, SIST, ShanghaiTech University.}
\thanks{The authors are with School of Information Science and Technology, ShanghaiTech University, Shanghai 201210, China
	and also with the MoE Key Laboratory of Intelligent Perception and Human-Machine Collaboration (ShanghaiTech University), Shanghai 201210, China
	(email:
		\href{mailto:xubw@shanghaitech.edu.cn}{xubw@shanghaitech.edu.cn}
		\href{mailto:zhaoxt@shanghaitech.edu.cn}{zhaoxt@shanghaitech.edu.cn}
		\href{mailto:fengdl@shanghaitech.edu.cn}{fengdl@shanghaitech.edu.cn}
		\href{mailto:yangyy2@shanghaitech.edu.cn}{yangyy2@shanghaitech.edu.cn}
		\href{mailto:soerensch@shanghaitech.edu.cn}{soerensch@shanghaitech.edu.cn}
	).}
}

\begin{document}

%
%


\marsPublishedIn{Submitted to:} 

\marsVenue{IEEE Transactions on Robotics}


\marsPlainAutors{Bowen Xu, Xiting Zhao, Delin Feng, Yuanyuan Yang, and S\"oren Schwertfeger}




\marsIEEE{}


\makeMARStitle

%
%

\maketitle

\begin{abstract}
This paper presents the ShanghaiTech Mapping Robot, a state-of-the-art unmanned ground vehicle (UGV) designed for collecting comprehensive multi-sensor datasets to support research in robotics, Simultaneous Localization and Mapping (SLAM), computer vision, and autonomous driving. The robot is equipped with a wide array of sensors including RGB cameras, RGB-D cameras, event-based cameras, IR cameras, LiDARs, mmWave radars, IMUs, ultrasonic range finders, and a GNSS RTK receiver. The sensor suite is integrated onto a specially designed mechanical structure with a centralized power system and a synchronization mechanism to ensure spatial and temporal alignment of the sensor data. A 16-node on-board computing cluster handles sensor control, data collection, and storage. We describe the hardware and software architecture of the robot in detail and discuss the calibration procedures for the various sensors and investigate the interference for LiDAR and RGB-D sensors. The capabilities of the platform are demonstrated through an extensive outdoor dataset collected in a diverse campus environment. Experiments with two LiDAR-based and two RGB-based SLAM approaches showcase the potential of the dataset to support development and benchmarking for robotics. To facilitate research, we make the dataset publicly available along with the associated robot sensor calibration data: 
\url{https://slam-hive.net/wiki/ShanghaiTech_Datasets}
\end{abstract}


\section{Introduction}

\begin{figure}[tbp]
	\centerline{\includegraphics[width=\linewidth]{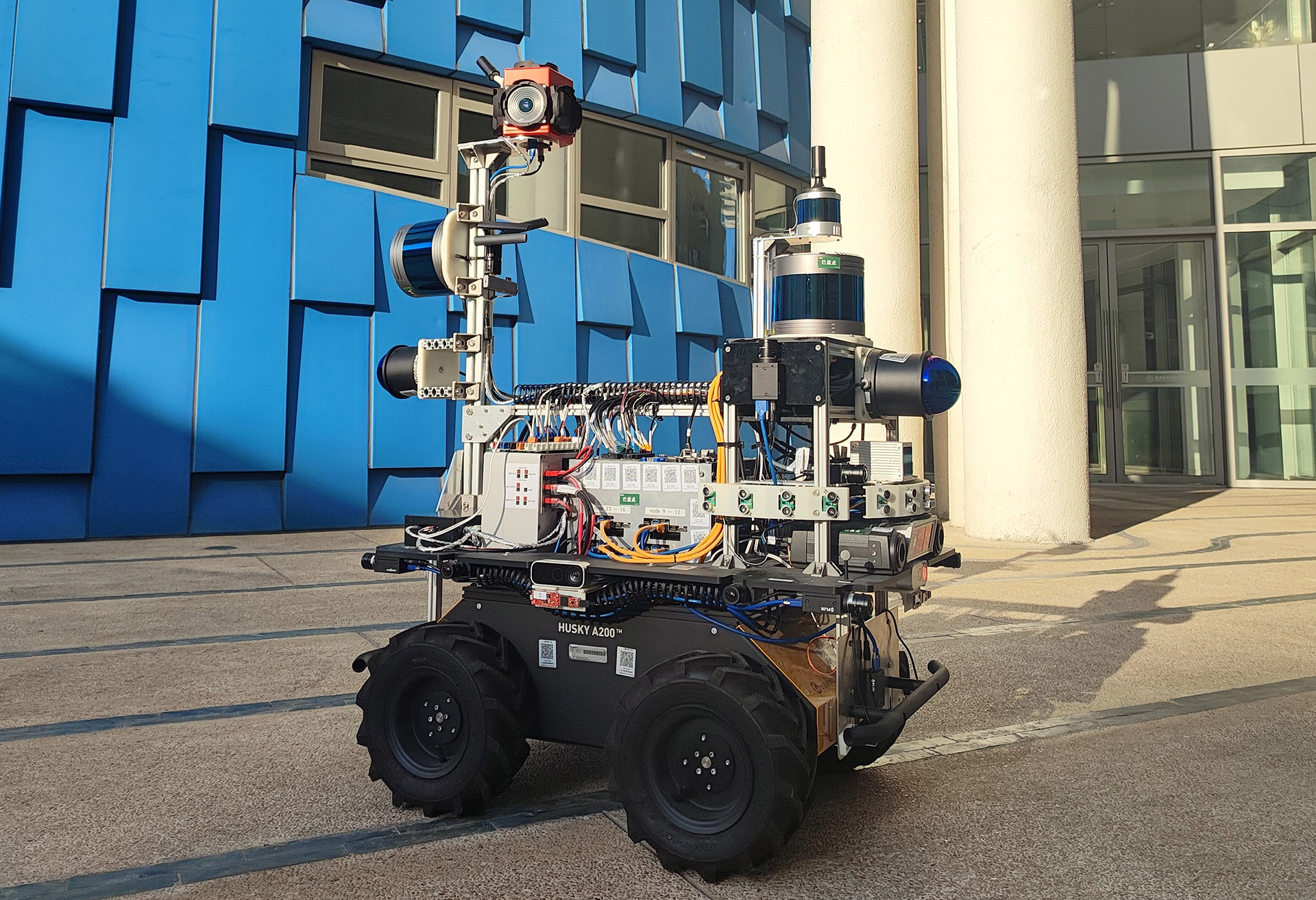}}
	\caption{The ShanghaiTech Mapping Robot.}
	\label{fig:mapping_robot_overview}
\end{figure}

Localization is an essential task of all mobile robots - they need to know their pose for path planning and navigation, as well as to perform their actual task. While there exist localization methods that use external infrastructure, such as GPS or special indoor markers \cite{apriltag2011}, the most general and flexible way of localization is to localize in a map \cite{dellaert1999monte}. Additionally, maps themselves are also needed for path planning and navigation. Thus, mapping is also a very important task for many mobile robotic systems. Typically, we need to solve those two problems together, because localization needs the map and mapping needs localization, with algorithms called Simultaneous Localization and Mapping (SLAM) \cite{cadena2016past}. Being such an essential algorithm, there is lots of research on SLAM. For proper scientific evaluation of the performance of SLAM algorithms, authors need to compare their solution to other, state-of-the-art approaches. For that, datasets of recorded robotic sensor data are used and then we typically compare the ground truth robot poses of these datasets with the poses estimated by the SLAM algorithms. 

One major problem with this approach is that there are various SLAM methods that use a multitude of different sensors and combination thereof, e.g. LiDAR, monocular camera, stereo, depth camera, event camera, radar, ultrasound, with or without IMU or odometry data. Those sensors may also be pointed in different directions or feature different resolutions or frame rates. In order to test and compare their solutions, SLAM researchers need to choose datasets compatible with their approach. But that means, they can only compare their algorithm with other methods that work with similar datasets. Often researchers even need to collect their own dataset, suitable for their SLAM approach. The consequence of all this is that currently most SLAM papers only compare to a very limited number of other state-of-the-art algorithms, and it is thus often unclear how well their solution performs compared to many other, state of the art methods. 

So ideally, there should be a dataset that can be used by all SLAM algorithms. Such a dataset would feature all kinds of different sensors in many different configurations, with very high specifications (e.g. resolution and frame-rate), such that SLAM algorithms can pick and choose which sensors they want, while potentially sub-sampling the data to fit their need. It is also not sufficient to do recording sessions with the different sensors in the same environment - what is needed for proper fair, repeatable and reproducible evaluation of the algorithms is, that all sensors should follow the exact same trajectory. To collect such a dataset we thus need a robot that can collect all this different sensor data in one go. In this paper we present our efforts to build such a robot: The ShanghaiTech Mapping Robot, which is all the mapping robot needed for ground vehicle dataset collection. 

The requirements for this dataset collection system are as follows: It should be a system for indoor and outdoor environments and as compact as possible. Because many sensors should be supported, which are in total quite heavy, flying systems, legged as well as hand-held solutions are impossible, so a wheeled mobile robotics base has to be chosen. While still being reasonable in terms of size, weight and consumption of power and computation resources, the robot should feature 1) as many different types sensors as possible with 2) as high specifications as possible and 3) with as big fields of view as possible. The ShanghaiTech Mapping Robot thus features 36 sensors of 9 different  modalities. The amount of data collected by all of these sensors can be up to 3,000 MiB per second. To facilitate the collection and storage of all this data we equipped our robot with a 16-node cluster. The robot is depicted in Fig.~\ref{fig:mapping_robot_overview}. 

This work was inspired by our first mapping dataset collection robot, which was much simpler and featured far fewer sensors \cite{chen2020advanced}. In \cite{9725992} we already reported on the design, construction and experiments of the 16-node "Cluster on Wheels". The ultimate goal of this work is to collect datasets that can be utilized by the SLAM Hive Benchmarking Suite~\cite{liu2024benchmarkingslamalgorithmscloud, yang2023slam} to properly evaluate SLAM algorithms by performing 10s of thousands mapping runs and analyzing their results in-depth. 

The core contributions of this paper are:
\begin{itemize}
	\item In-depth discussion of the requirements, design and construction of the ultimate dataset collection robot;
        \begin{itemize}
        \item Mechanical Design
        \item Electrical System
        \item Sensor Setup
        \item Computing System
        \item Distributed Software System
        \item Hardware-Synchronization
        \end{itemize} 
    \item Calibration approach, especially extrinsic calibration of 32 sensors
    \item Sensor Interference Evaluation
	\item Outdoor Dataset\footnote{The dataset is available at \url{https://slam-hive.net/wiki/ShanghaiTech_Datasets}} with SLAM Experiments
\end{itemize}

The rest of this paper is organized as follows: Section \ref{sec:related} reviews related work on SLAM datasets and multi-sensor robotic platforms. Section \ref{sec:sensors} describes the sensors of our data collection system, while Section \ref{sec:backbone} discusses the basic backbone systems of the robot and Section \ref{sec:integration} introduces the hardware and software integration of the sensors with the cluster as well as the hardware synchronization system. The intrinsic and extrinsic calibration procedures of the sensors are explained in Section \ref{sec:calibration}. Section \ref{sec:procedures} then reports on the data collection and post-processing procedures. A few sensors interfere with each other and the limited effect of this is investigated and described in Section \ref{sec:Interference}. The outdoor dataset provided with the paper, as well as SLAM experiments with it, are presented in Section \ref{sec:outdoor}. The opportunities and future work that our robot suggests are discussed in Section \ref{sec:futurework} and this paper concludes with Section \ref{sec:conclusions}. Details of the data collected by the robot are provided in Appendix \ref{appendix:dataset_topics}.




\section{Related Works}
\label{sec:related}

The development of acquisition of SLAM datasets has been crucial for advancing research in robotics, computer vision, and autonomous driving. This section reviews relevant datasets and techniques, focusing on multi-sensor approaches, visual SLAM, LiDAR-based systems, event-based cameras, and mmWave radar applications.


Several large-scale datasets have been instrumental in driving progress in autonomous vehicle research. The KITTI dataset \cite{10.1177/0278364913491297}, introduced in 2013, was one of the pioneering efforts, providing synchronized stereo imagery, LiDAR scans, and GPS/IMU data. Building upon this foundation, the Oxford RobotCar dataset \cite{doi:10.1177/0278364916679498} offered long-term data collection over varying conditions, while the Cityscapes dataset \cite{7780719} focused on semantic understanding of urban scenes. More recent contributions like nuScenes \cite{9156412} and Waymo Open Dataset \cite{9156973} have further expanded the scope, incorporating additional sensor modalities and more diverse driving scenarios. The Ford Multi-AV Seasonal Dataset \cite{doi:10.1177/0278364920961451} specifically addresses the challenges of seasonal variations. Our work distinguishes itself from these datasets by offering a more extensive and diverse sensor suite, including event-based cameras, infra-red cameras, a combination of 5 different LiDARs models, and a multi-directional mmWave radars setup, which are not commonly found in existing datasets. Furthermore, we do support the collection of indoor, which is not possible with passenger vehicles.


LiDAR approaches are one of the prevalent SLAM methods, as well as for 3D reconstruction and autonomous driving applications. Datasets such as M2DGR \cite{9664374} and the Multi-Modal LiDAR Dataset \cite{9981078} have provided valuable resources for developing and evaluating LiDAR-based SLAM algorithms. The RUGD dataset \cite{8968283} specifically targets unstructured environments, while PandaSet \cite{9565009} offers a comprehensive sensor suite including LiDAR for autonomous driving research. Our work builds upon these efforts by integrating high-resolution LiDAR data with other sensor modalities, allowing for more robust and versatile perception algorithms.


Visual SLAM, as yet another key approach of SLAM, has seen significant advancements, fundamentally owing to the availability of diverse datasets. The EuRoC MAV datasets \cite{doi:10.1177/0278364915620033} provided benchmark data for aerial vehicles, while more recent works like the multi-camera framework by Kaveti et al. \cite{10253964} and the visual-inertial odometry system by Zhang et al. \cite{9662207} have explored multi-camera setups for improved robustness. The Campus Map dataset \cite{campus_map} offers a large-scale multi-camera dataset with additional sensors, similar to our approach. However, our work extends beyond this by incorporating event-based cameras and mmWave radars, enabling research into fusion algorithms that can leverage these novel sensing modalities.


Event-based cameras have emerged as a promising technology for high-speed and high-dynamic-range applications. Datasets such as ViViD++ \cite{9760091}, DSEC \cite{9387069}, and VECtor \cite{9809788} have provided valuable resources for developing event-based SLAM and odometry algorithms. The Event-Camera Dataset and Simulator \cite{doi:10.1177/0278364917691115} and the Multivehicle Stereo Event Camera Dataset \cite{8288670} have further expanded the available data for event-based research. Our work incorporates event-based cameras alongside traditional sensors. It allows for the development of hybrid algorithms like \cite{8258997} that can exploit the complementary strengths of different sensor modalities while also creates a benchmark for various event-based algorithms to compete with their visual or LiDAR-based counterparts.


The use of mmWave radar in SLAM and autonomous driving is an emerging field with significant potential. Recent works such as the multi-radar inertial odometry by Huang et al. \cite{huang2023multi} and the 4D imaging radar inertial odometry and mapping (iRIOM) by Zhuang et al. \cite{10100861} have demonstrated the capabilities of mmWave radar for ego-motion estimation and mapping. Park et al. \cite{9495184} proposed a novel radar velocity factor for pose-graph SLAM, while Lu et al. \cite{10.1145/3386901.3388945} developed milliMap for robust indoor mapping using mmWave radar. Our robot includes mmWave radar data, enabling research into multi-sensor fusion algorithms that can leverage the unique properties of radar, such as detecting glass, using the Doppler effect to measure object's speed and its robustness in adverse weather conditions.

In summary, while existing datasets have made significant contributions to various aspects of SLAM and autonomous driving research, our work uniquely combines a wide array of sensor modalities, including event-based cameras and mmWave radars. This comprehensive approach allows for the development and evaluation of more robust and versatile algorithms capable of operating in diverse and challenging environments.

\section{Sensor Suite}
\label{sec:sensors}

Ground vehicles like road automotives, warehouse AGVs, or even household robot vacuum cleaners, all have different sensors equipped, suiting their size, environment of operation and budget. Moreover, the choice of sensor is affected by the indented mission of the vehicle. Most vehicles would use the data for basic localization and mapping, while some of them also expect fine reconstruction of environment or detecting surrounding objects they try to avoid bumping into (or contrarily, would like to interact with).

Different types of sensors suit different tasks. To achieve the goal of collecting universal ground vehicle datasets, we tried our best to equip our robot with almost all kinds of sensors utilized in the SLAM and self-driving field, hoping that the datasets it produces would fit most, if not all, SLAM algorithms. To be specific, our choice covers visual sensors (including frame-based RGB cameras, depth cameras, infra-red cameras and event-based cameras), LiDARs (including rotational ones and solid ones), inertial sensors, mmWave radars, ultrasonic range finders and GNSS receivers. In this section, we will introduce the sensor suite on the robot.

\begin{figure*}[htbp]
	\centerline{\includegraphics[width=\linewidth]{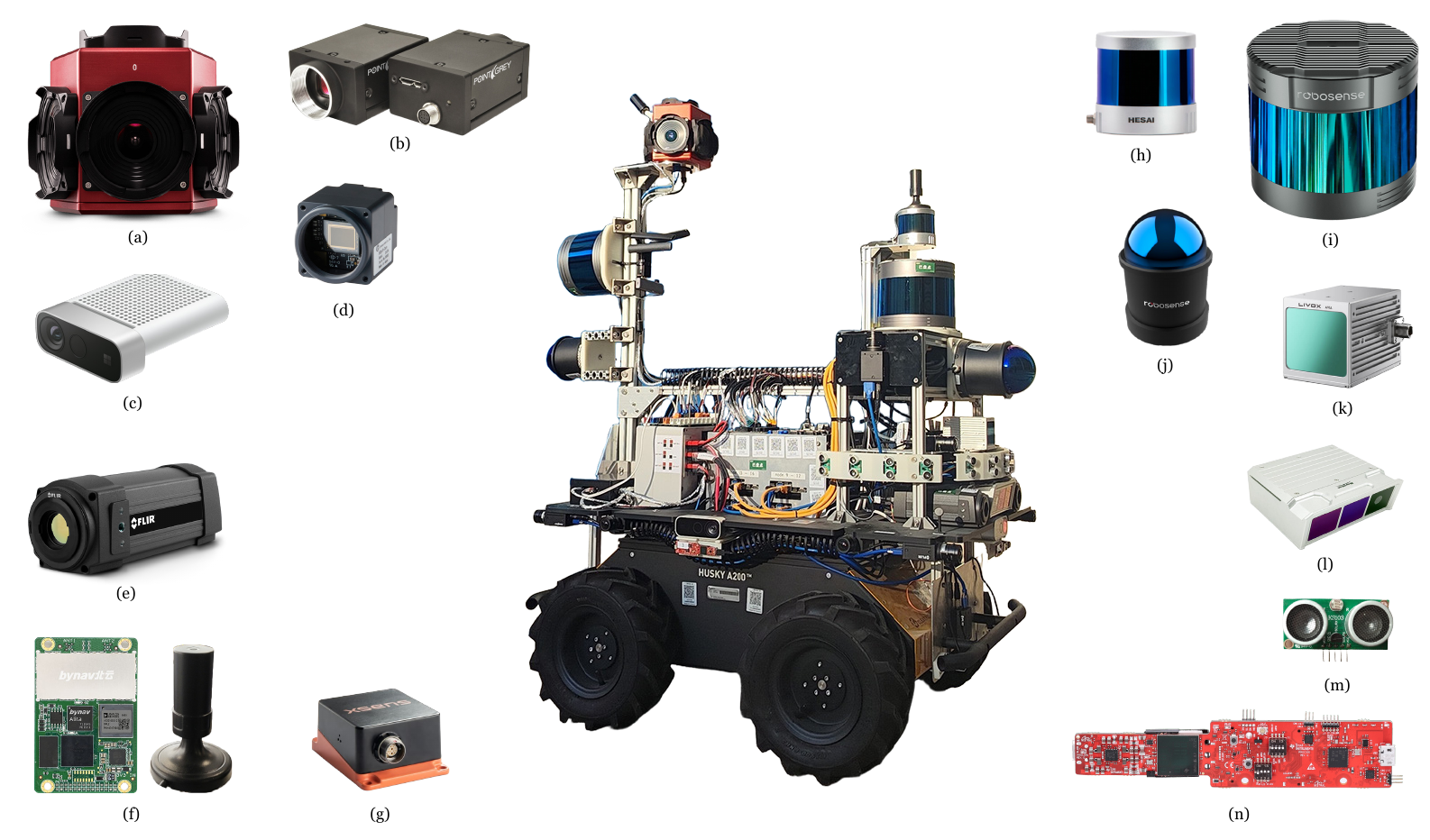}}
	\caption{Sensors on ShanghaiTech Mapping Robot.}
	\label{fig:sensors_overview}
\end{figure*}

\begin{table*}[htbp]
\centering
\caption{Sensor Configuration of the ShanghaiTech Mapping Robot}
\label{tab:sensor_config}
\resizebox{\textwidth}{!}{%
\begin{tabular}{l|l|c|l|c|l|l|l}
\hline
\textbf{Sensor Type} & \textbf{Model} & \textbf{Qty} & \textbf{Specifications} & \textbf{Max Data Rate} & \textbf{Field of View} & \textbf{Looking Direction} & \textbf{Appearance} \\
\hline
RGB Camera & FLIR Grasshopper3 & 11 & 2448 x 2048 pixels & 75 fps & 81.9° x 61.2° & Front, Back, Left, Right, Up (stereo); Down (mono) & Fig. \ref{fig:sensors_overview}-b\\
RGB Camera & FLIR LadyBug5+ & 1 (6) & 6 x 2464 x 2048 pixels & 30 fps & 90\% of full sphere & Omnidirectional & Fig. \ref{fig:sensors_overview}-a\\
RGB-D Camera & Azure Kinect DK & 5 & RGB: 3072 x 1920, Depth: 1MP & 30 fps & N/A & Front, Back, Left, Right, Up & Fig. \ref{fig:sensors_overview}-c\\
IR Camera & FLIR A315 & 2 & 320 x 240 pixels & 60 Hz & 25° & Front (stereo pair) & Fig. \ref{fig:sensors_overview}-e\\
Event-based Camera & EvC3A & 2 & 640 x 480 pixels & N/A & N/A & Front (stereo pair) & Fig. \ref{fig:sensors_overview}-d\\
Spinning LiDAR & Robosense Ruby & 2 & 128 beams, 200m range & 10 Hz & 360° x 40° & Horizontal, Vertical (90° pitched) & Fig. \ref{fig:sensors_overview}-i\\
Spinning LiDAR & Hesai PandarQT & 1 & 64 beams, 60m range & 10 Hz & 360° x 104.2° & Horizontal & Fig. \ref{fig:sensors_overview}-h\\
Spinning LiDAR & Robosense Bpearl & 2 & 32 beams, 30m range & 10 Hz & 360° x 90° & Front, Back & Fig. \ref{fig:sensors_overview}-j\\
Solid-state LiDAR & Livox Avia & 1 & 190m range & 10 Hz & 70.4° x 77.2° & Front & Fig. \ref{fig:sensors_overview}-k\\
Solid-state LiDAR & Neuvition M1 & 1 & 200m range & 10 Hz & 45° x 25° & Front & Fig. \ref{fig:sensors_overview}-l\\
IMU & Xsens MTi-630R & 1 & 400 Hz update rate & 400 Hz & N/A & N/A & Fig. \ref{fig:sensors_overview}-g\\
Ultrasonic Sensor & KS103 bundle& 1 & 1cm to 8m range & N/A & 13 different angles & Front & Fig. \ref{fig:sensors_overview}-m\\
mmWave Radar & TI IWR6843AOPEVM & 5 & 60GHz & N/A & 120° x 120° & Front, Back, Left, Right, Up & Fig. \ref{fig:sensors_overview}-n\\
GNSS Receiver & Bynav A1 & 1 & Dual antenna with IMU & N/A & N/A & N/A & Fig. \ref{fig:sensors_overview}-f\\
\hline
\multicolumn{2}{r|}{\textbf{Total}} & \textbf{36 (41)} & \multicolumn{4}{l}{\textbf{17 RGB, 5 RGB-D, 2 Event-based, 2 IR, 7 LiDARs, 5 mmWave Radars, 1 Sonar Set, 1 IMU, 1 GNSS RTK/INS}} \\
\hline
\end{tabular}%
}
\end{table*}

\subsection{RGB Cameras}
\label{subsection:gs3}

Color cameras are some of the most used sensors in the SLAM field, providing data for visual SLAM algorithms, as well as providing input for advanced computer vision algorithms like semantic segmentation or object detection. Conventional algorithms such as ORB-SLAM3 utilize either monocular or stereo visual streams \cite{ORB3}.  


Bird's-Eye-View data, stitched from omni-directional cameras, is also very popular recently, e.g. \cite{li2022bevformer}.


Thus we equipped the mapping robot with 2 models of RGB cameras, FLIR Grasshopper3 (GS3-U3-51S5C-C) and FLIR LadyBug5+.

FLIR Grasshopper3 (GS3-U3-51S5C-C) 
(Fig. \ref{fig:sensors_overview}-b)
is a 2/3-inch color camera that has a resolution of $2448 \times 2048$ pixels (5MP) and can capture images in global shutter mode at a frame rate of up to 75fps.\cite{gs3}
Our incentive was to use them to achieve stereo views towards five main directions: front, back, left, right, up, and to have a monocular downward view of the floor or ground the robot drives on.
FLIR LadyBug5+ 
(Fig. \ref{fig:sensors_overview}-a)
is a polydioptric omnidirectional color camera, consisting of 6 pinhole cameras. 5 of the cameras face the surrounding while the last one faces upward. Each of the 6 sensors has a resolution of $2464 \times 2048$ and a maximum frame rate of 30fps \cite{ladybug5p}.


\subsection{RGB-Depth Cameras}

Compared to RGB cameras, which provide no depth measurements, RGB-Depth cameras capture both color (RGB) and depth information simultaneously. They combine a standard color camera with a depth sensor, providing a 3D representation of the captured scene.

We chose the Azure Kinect DK
(Fig. \ref{fig:sensors_overview}-c)
as the RGB-D camera of our robot. It features a 12MP high-resolution ($3072 \times 1920$) RGB camera that can capture color images in global shutter mode at 30fps and an 1MP depth camera that uses ToF technology to measure distances within the range of 0.5 - 5.46 meters \cite{azurekinect}. Compared to depth cameras that use structured light technology, ToF RGB-D cameras work better in outdoor environments and have a longer range, which is more suitable for our robot.

In order to cover the most field of view, we have mounted five Azure Kinect DKs, covering the front, back, left, right, and up directions.


\subsection{IR Cameras}

Far infrared (IR) cameras, more precisely thermal cameras, can measure temperature and thus be used to tell heat-emitting objects like human beings, animals, vehicles, etc., from the background. They are also good at detecting fire or malfunctioning electricity equipment.

For our robot, we chose FLIR A315
(Fig. \ref{fig:sensors_overview}-e), a compact uncooled microbolometer thermal camera. A315 features a $320 \times 240$ pixels Vanadium Oxide detector, having a temperature detection range from -20°C to 120°C with an accuracy up to ±2°C. Its built-in lens provides a 25° field of view and motorized focusing control. It has a spatial resolution of 1.36 mrad and a temperature resolution of 50 mK. Two A315s are used to form a forward looking stereo pair.

\subsection{Event-based Cameras}
\label{subsection:evc3a}

Unlike conventional frame-based cameras, event-based cameras capture asynchronous 'Contrast Detector' events representing per-pixel brightness changes and report them at real-time, instead of reading out full frames at a fixed rate, which means that event-based cameras can distinguish motion at a very high speed rate, introducing a new approach into the field of visual SLAM.

We equipped the robot with EvC3A event-based cameras
(Fig. \ref{fig:sensors_overview}-d)
featuring the Prophesee PPS3MVCD sensor (often be referred as the Gen3.1 VGA Sensor) also used in the EVK3 camera \cite{evc3a}. It has a resolution of $640\times480$ on its 3/4” sensor, achieving a typical latency of 200µs and a dynamic range above 120dB. Two of them are deployed to form a forward looking stereo pair.

\subsection{LiDARs}

LiDAR-based approaches have always, along with their visual counterparts, sat in the center of the field of SLAM. Featuring straightforward capability in range measuring, LiDARs appear on the whole spectrum of ground vehicles, from \$500 robot vacuum cleaners to \$50,000 road automotives. To produce a universal ground vehicle dataset, LiDARs, many of them, are a must.

In order to diversify the technical methodologies employed by the LiDAR systems and facilitate comprehensive coverage of the sensing area in all directions, we equipped the robot with seven LiDARs, encompassing five distinct models: Robosense Ruby, Robosense Bpearl, Hesai PandarQT, Livox Avia,
and Neuvition M1. 


Robosense Ruby and Hesai PandarQT are conventional rotational LiDARs. Robosense Ruby
(Fig. \ref{fig:sensors_overview}-i)
is a 128-beam LiDAR with a 0.1° angular resolution on both the horizontal and vertical axes, 200m range and 40° vertical field of view, well-suited for autonomous driving applications requiring far detection range. In contrast, Hesai PandarQT
(Fig. \ref{fig:sensors_overview}-h)
, also known as QT64, is a 64-beam LiDAR designed for indoor applications. It features a 60m range, 0.6° horizontal angular resolution and has a vertical angular resolution up to 1.45°. With a 104.2° (-52.1° to +52.1°) vertical aperture, its FOV covers a typical room from the ceiling to the floor. Both of them, as well as all other LiDARs equipped by our robot, run in dual-return or triple-return mode, which can improve reflection detection accuracy \cite{zhao2020mapping}\cite{zhao20243dref}\cite{li2024detection}.

Robosense Bpearl
(Fig. \ref{fig:sensors_overview}-j)
is a 32-beam rotational LiDAR featuring a hemispherical $360^{\circ}\times90^{\circ}$ field of view. It has a small blind range of less than 0.1m and a maximum range of 30m, designed for near-field detection. It provides a vertical resolution of 2.81° and a horizontal resolution up to 0.2°.


Livox Avia and Neuvition M1 are unconventional LiDARs. Livox Avia
(Fig. \ref{fig:sensors_overview}-k)
is a semi-solid-state LiDAR featuring a unique non-repetitive scanning pattern and a forward-looking field of view. Unlike conventional rotational LiDARs, Livox Avia's coverage area ratio increases with the time of integration. It has a $70.4^{\circ}\times77.2^{\circ}$ field of view, 190m range and ±2cm accuracy. It features a best point density among all 5 LiDARs equipped by our robot and thus fits use cases like fine mapping and 3D reconstruction. Neuvition M1
(Fig. \ref{fig:sensors_overview}-l)
is a MEMS-based solid-state LiDAR, providing a $45^{\circ}\times25^{\circ}$ field of view with a $1750\times480$ resolution and 200m range at 20\% reflectivity.

\subsection{IMUs}

Inertial sensors are proprioceptive sensors that do not rely on signals returned from objects and thus are most environment-adaptive among all kinds of sensors (geomagnetic and gravitational fields do help but are not necessity, even though they are stable at many places on Earth). They can either be used alone for dead reckoning or work with cameras, LiDARs or GNSS receiver to perform multi-sensor fusion localization.

We chose the MTi-630R
(Fig. \ref{fig:sensors_overview}-g), a rugged and compact attitude and heading reference system (AHRS) from Xsens, to be the main Inertial Measurement Unit (IMU) of our robot. It provides 0.2° RMS roll/pitch and 1° RMS heading accuracy and is able to perform high precision strapdown integration at 400Hz \cite{mti630r}. It also measures temperature and atmospheric pressure, potentially helping to determine the elevation of robot.

Devices like Livox Avia, Azure Kinect DKs, and the GNSS receiver also have their own built-in IMU.

\subsection{Ultrasound Range Finders}

Ultrasound Range Finders are often equipped by road vehicles for detecting near-field obstacles, especially in the context of automatic parking. We equipped our robot with 13 KS103 ultrasonic range finders
(Fig. \ref{fig:sensors_overview}-m). KS103 has a range of 1cm to 8m and a resolution of 1cm. It utilizes proprietary filtering techniques to improve ranging accuracy, claiming high precision even in noisy environments.


\subsection{mmWave Radars}

mmWave Radar is yet another type of sensor that often appears on road vehicles. They can robustly detect the position and relative speed of objects like pedestrians, cyclists, and other vehicles, and are thus often used to provide functions like Blind Spot Detection and Autonomous Emergency Braking.


For our robot, we chose IWR6843AOPEVM
(Fig. \ref{fig:sensors_overview}-n), a 60GHz mmWave radar from Texas Instruments (TI). It achieves a wide 120° azimuth and 120° elevation field of view with a small antenna-on-package module \cite{iwr6843aopevm}. We decided to have five of these radars to look at front, back, left, right, and up directions.

\subsection{GNSS receiver}

To measure an accurate trajectory serving as ground truth in outdoor use cases, Global Navigation Satellite System (GNSS) is the best choice. With the assistance of differential reference stations, onboard IMU, and some magic of post-mission processing, GNSS receivers can achieve an accuracy up to 1cm.

We employed a Bynav A1 GNSS RTK (Real Time Kinematic) receiver
(Fig. \ref{fig:sensors_overview}-f) that utilizes GPS, GLONASS, Galileo, BeiDou, QZSS, and NAVIC constellations. The receiver supports dual antenna setup, measuring not only the position but also the attitude of the robot. It also has a builtin Inertial Navigation System (INS) that helps improve positioning both in real-time and for processing afterwards.

\section{System Backbone}
\label{sec:backbone}

To make all the sensors mentioned above work simultaneously on our robot, proper data communication, synchronization, powering, as well as a moving base are essential. In this section, we will introduce several main parts of the system hardware acting as the backbone infrastructure for our sensing platform.

\subsection{Husky UGV}

Our robot is built upon the Clearpath Husky, a 990mm long, 670mm wide UGV (Unmanned Ground Vehicle) equipped with rugged construction, high-torque differential drive and lug-tread tires. It has full terrain mobility that allows us to collect datasets in a whole span of various environments, from the interior of buildings, paved roads, to grass fields or even forests.

\begin{figure}[htbp]
	\centerline{\includegraphics[width=0.9\linewidth]{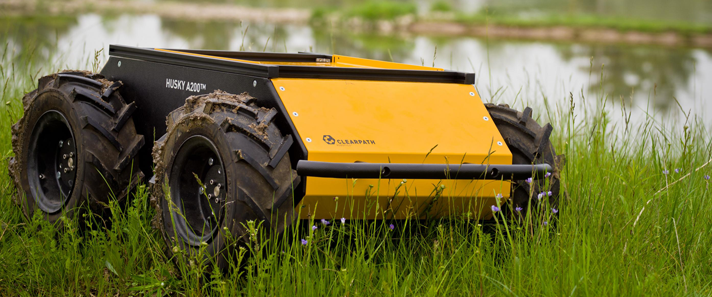}}
	\caption{The Clearpath Husky UGV.}
	\label{fig:husky}
\end{figure}

\subsection{Computation Cluster}
\label{subsection:cluster}

To control all the sensors and record the data captured by them, a powerful computer is essential for our mapping robot.
Since there are too many sensors on the mapping robot, each of them capturing massive amounts of data, even the most powerful portable computer does not have enough I/O and storage bandwidth to handle the recording of the data from all the sensors.
Thus a very compact, 16-node computation cluster was specially designed for this usage, whose development has been described in \cite{9725992}.

\begin{figure}[htbp]
	\centerline{\includegraphics[width=\linewidth]{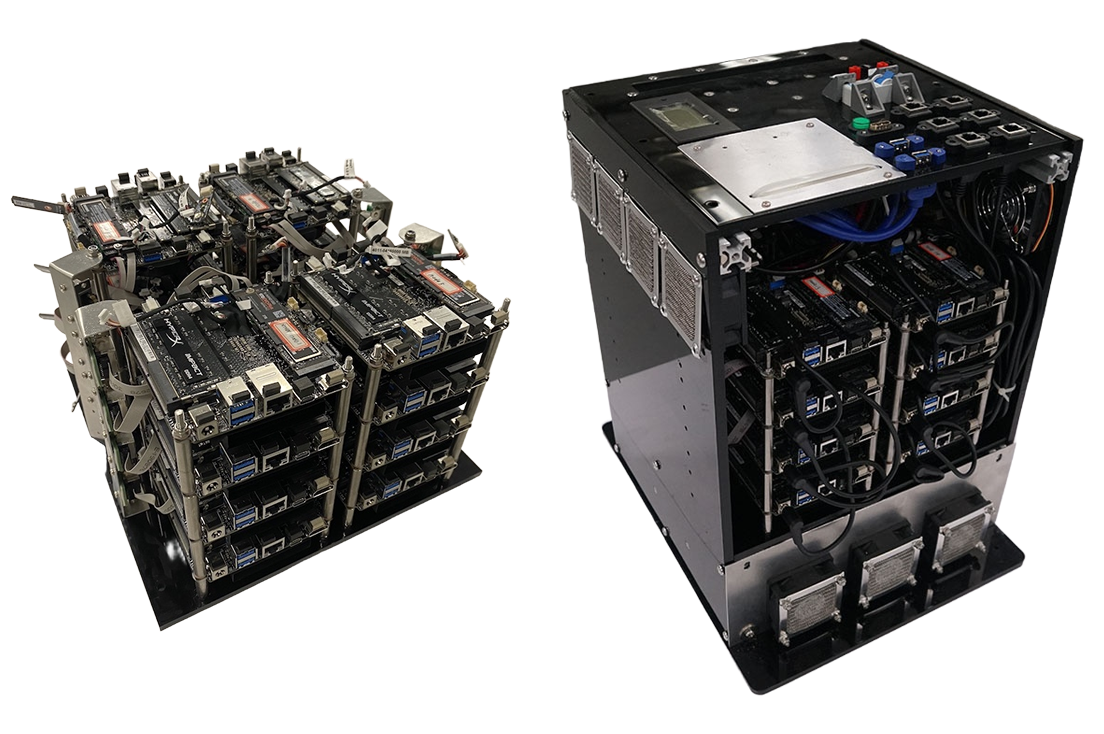}}
	\caption{The 16 nodes of the cluster (left) and the complete cluster without one side wall (right).}
	\label{fig:cluster}
\end{figure}

The cluster consists of 16 computing nodes, each modified from an \textit{ASUS Mini PN51} computer, having an 8-core, 16-threads \textit{Ryzen 7 5700U} processor,  32 GiB of RAM, 2 TiB of NVME M.2 SSD storage (Samsung 970EVO Plus, sequential write speed up to 3,300 MB/s), 2 Gigabit Ethernet ports (one on the main board, one in an M2 slot, replacing the WiFi), 3 USB 3.2 Gen1 Type-A ports and 2 USB 3.2 Gen2 Type-C ports.
The nodes are stacked in a customized case which also accommodates several DC-DC converters and a LAN switch providing power and interconnection for the cluster, as shown in Fig. \ref{fig:cluster}.

Among the nodes there is one master node that acts as the controller of the whole cluster, while the other worker nodes do not have their operating system installed locally but utilize network boot and use the system image hosted on the master node, making it much easier to configure the sensor drivers and orchestration programs over all 16 nodes.
This approach also saves storage on worker nodes, allowing a longer record time.

Each worker node is intended to control and record up to 4 sensors, including 3 through USB 3.0 connections and one through a Gigabit Ethernet connection.
Sensor drivers help configure the sensors connected to the according node and collect the captured data.
Raw sensor data will be saved in situ to the SSD mounted on the worker node and can, on demand, be streamed to the master node through ROS.
Recorded sensor data can be collected for post-processing by either copying them to external hard drives connected to worker nodes or forwarding them via Gigabit Ethernet through the master node to an external server.

\subsection{Synchronization Board}
\label{subsection:sync}

To eliminate or remedy the time difference of data collected by different sensors, both electronic synchronization and IEEE 1588 PTP (Precision Time Protocol) are the most preferred approaches. Both of them can achieve sub-microsecond precision. We developed Syncboard, a combination of a Raspberry Pi 4 computer and a specially designed I/O shield, to act as a master clock of the whole robot that provides triggering or the time base for the sensors equipped as well as the computation nodes in the cluster. On the Syncboard there are 12 triggering output channels and a LiDAR synchronization channel, capable of connecting at most 130 sensors. 

\begin{figure}[htbp]
	\centerline{\includegraphics[width=\linewidth]{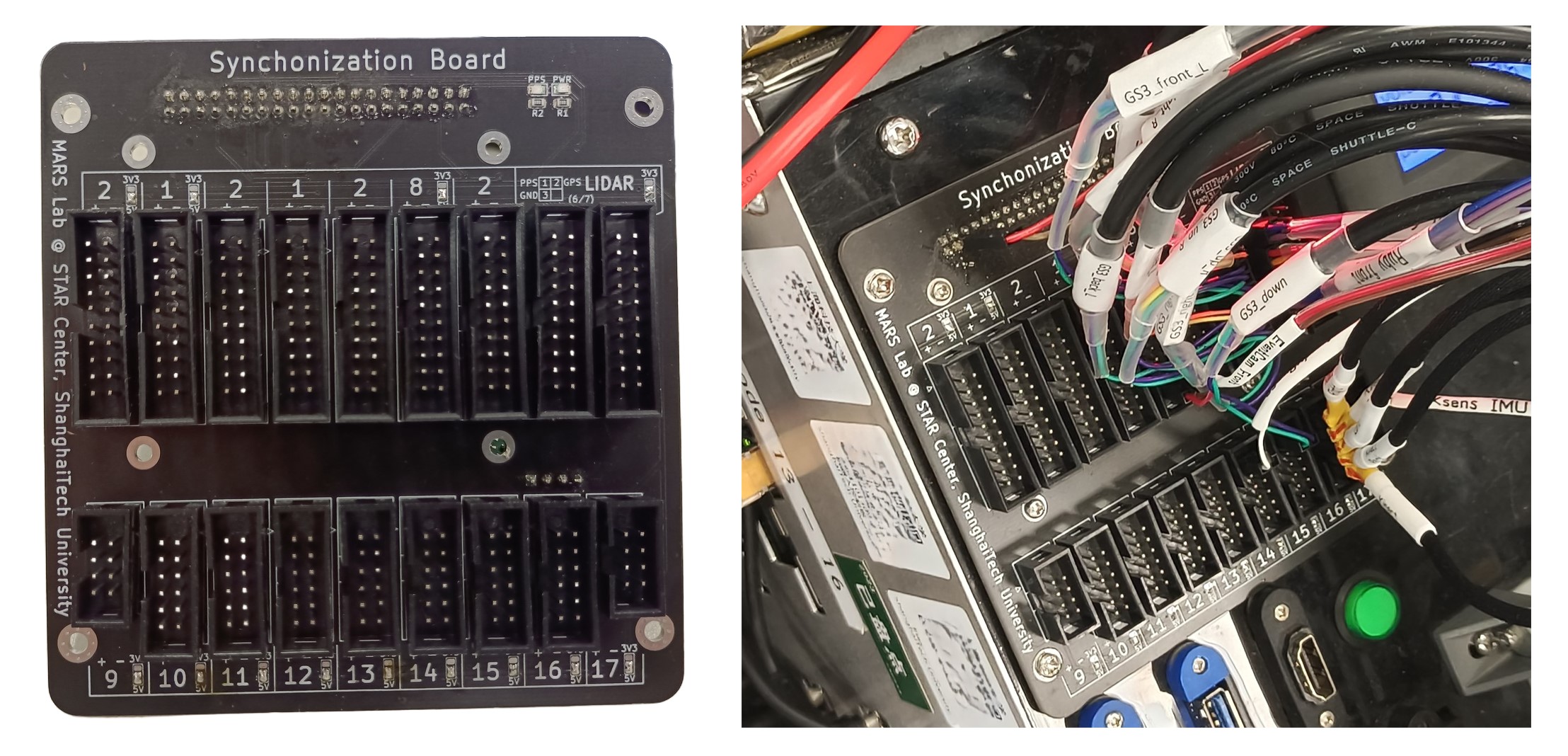}}
	\caption{The front side of Syncboard (left). Syncboard installed on the cluster and plugged with triggering cables of sensors (right). }
	\label{fig:sync_plugged_and_not}
\end{figure}

Each of the 12 triggering output channels can generate square wave signals according to the channel-wise designated parameters including frequency (up to 1000Hz), triggering edge polarity (rising edge, failing edge or both edges), time offset and duty ratio, adapting to various triggering-compatible sensors like cameras and IMU. The voltage level of each channel can be set to 3.3V or 5V by changing the soldering bridge on the board. One channel may have 6 to 40 vacancies available for sensor connection, enabling us to easily trigger multiple sensors with the same signal.

The LiDAR synchronization channel, on the other hand, is capable of generating GPRMC+PPS signal. GPRMC, or the Recommended Minimum Specific GPS/Transit Data, is an NMEA 0183 message type designed for carrying positioning data, including a timestamp accurate to the second. It is often given along with a PPS (Pulse-per-Second) signal that indicates the exact beginning of the second contained in the message. Conventional LiDARs consume the GPRMC+PPS signal, often from a real GPS receiver in old days, to set their internal clock. The channel can be configured to have a certain baud rate as well as an original or inverted voltage level to adapt to different LiDARs. The channel provides 10 vacancies for LiDAR connection.

Additionally, Syncboard broadcasts its clock using PTP throughout the cluster LAN such that all computing nodes can synchronize themselves, as well as the sensors connected to them. Syncboard itself takes either GNSS or public NTP server as its external master clock.

Syncboard also provides a handy LED button for us to physically start or stop triggering and check its working status, no need to send commands on remote devices. Syncboard itself is integrated with the computation cluster. Its I/O shield acts as a part of cluster's top I/O plate and the Raspberry Pi is accommodated inside the cluster and connected to the Ethernet switch as the 17th node.

\subsection{Power Box}
\label{sec:powerbox}

We designed a power box to provide electricity to all sensors of various types as well as the computation cluster and the UGV base.
The power box is able to get power from lithium batteries, convert them into different voltage rails and distributes them to all electric devices.

The power box is equipped with 2 independent battery connectors. High current diodes are employed to isolate the two inputs, enabling the hot swap of batteries and preventing them from charging each other.
After flowing through the diodes, electricity meets the main breaker which protects the robot from over-current and acts as the power switch of the robot, then a digital power meter which displays the real-time voltage, current, power, and an integration of energy consumption, before being distributed into converters.

\begin{figure}[htbp]
	\centerline{\includegraphics[width=0.9\linewidth]{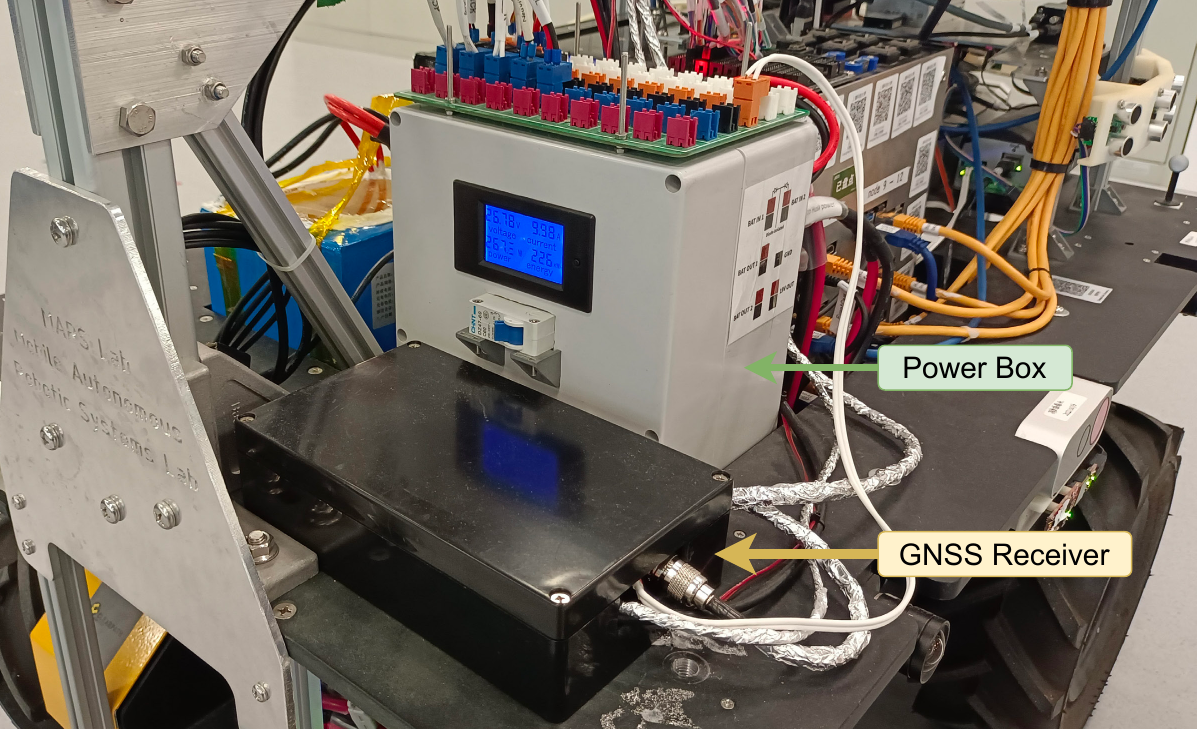}}
	\caption{The power box and GNSS receiver of ShanghaiTech Mapping Robot.}
	\label{fig:powerbox_and_gps}
\end{figure}

\begin{figure}[htbp]
	\centerline{\includegraphics[width=0.9\linewidth]{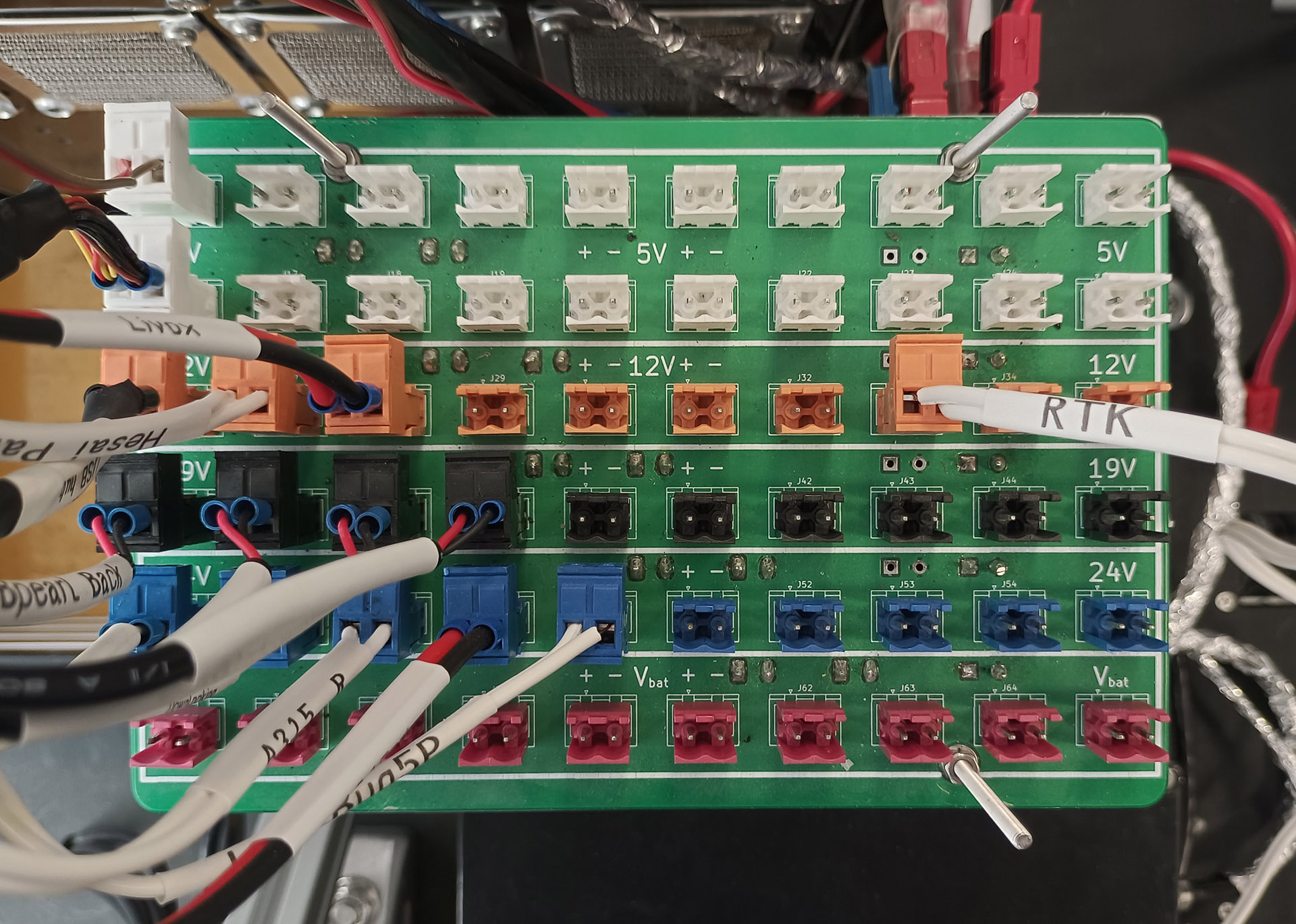}}
	\caption{The distribution board at the top of power box. Power lines of several sensors are connected to the box to get electricity of different voltage.}
	\label{fig:powerboard}
\end{figure}

Inside the power box, multiple DC-DC converters turn the battery power rail into 4 stable DC rails, 5V, 12V, 19V, 24V respectively.
The 4 converted rails along side with the battery power rail are connected to the power distribution board, shown in Fig. \ref{fig:powerboard}, at the top of the power box. Sensors get electricity from sockets on this board unless they use a single USB port for both data and powering. In that case, they get power from cluster nodes. The UGV base and the computation cluster have their power line connected to the back of the power box to get battery power rail directly.

\section{System Integration \& Sensor Setup}
\label{sec:integration}

With all sensors and backbone hardware ready, in this section we will introduce how we integrated all components together as well as the setup of each sensor.

\begin{figure}[htbp]
	\centerline{\includegraphics[width=\linewidth]{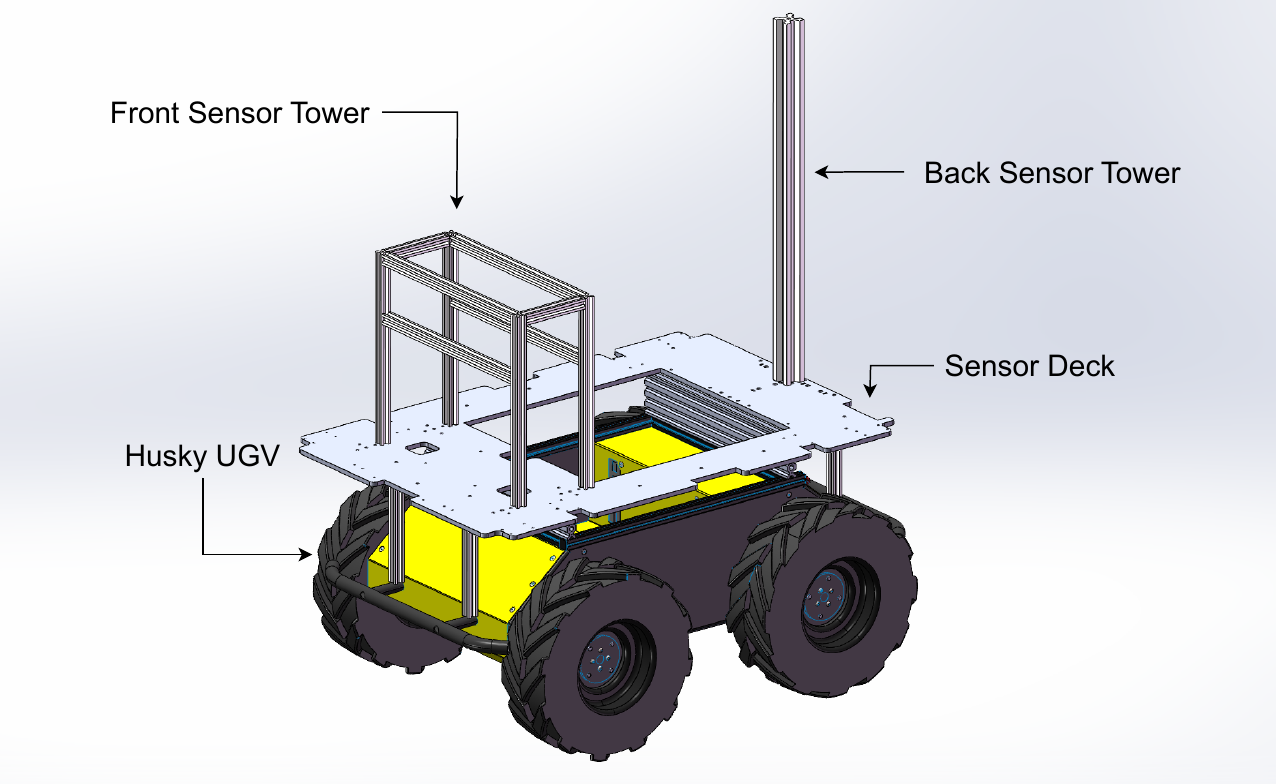}}
	\caption{A simplified model of ShanghaiTech Mapping Robot 's mechanical design, showing Clearpath Husky UGV, the sensor deck, front sensor tower and back sensor tower.}
	\label{fig:simple_model}
\end{figure}

The end goal for the robot is to collect very accurate sensor data. For this we will calibrate the extrinsics (the poses) of all sensors, which is described in Section \ref{sec:calibration}. To facilitate this, the sensors need to be rigidly mounted in such a way, that they do not move even when the base is driving over harsh terrain. We thus mechanically designed a very stiff mounting structure. As shown in Fig. \ref{fig:simple_model}, the robot features a sensor deck, a front sensor tower and a back sensor tower to accommodate all sensors. The back tower is a single 80 cm tall beam. This beam is screwed into the sensor deck, also with triangle brackets, secured by a back plate to minimize roll rotation and further stiffened by a diagonal aluminum profile connected to the sensor deck as well as a horizontal aluminium profile connected to the front tower to reduce pitch rotation. Fig.~\ref{fig:powerbox_and_gps} shows these support structures. The 60 cm tall front tower is a four-legged aluminum profile structure. 

\subsection{Front Sensor Tower}

The front sensor tower consists of multiple floors, each of them is made of either aluminum or acrylic plastic boards. 

On the top  mounts the front GNSS antenna as well as 2 rotational LiDARs, Hesai PandarQT and the first (front) Robosense Ruby.

\begin{figure}[htbp]
    \centerline{\includegraphics[width=\linewidth]{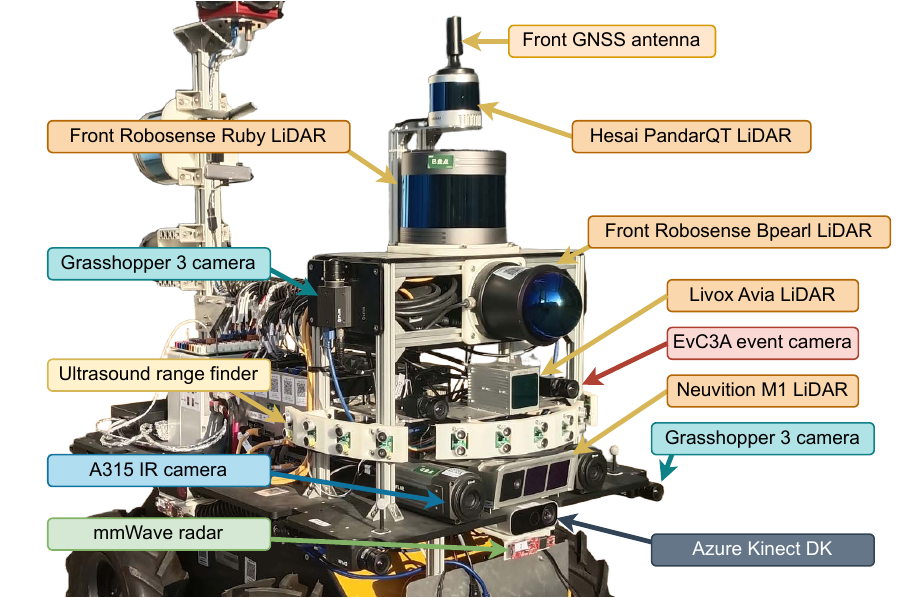}}
    \caption{The front sensor tower of ShanghaiTech Mapping Robot.}
    \label{fig:front_sensor_tower}
\end{figure}

As shown in Fig. \ref{fig:front_sensor_tower} and Fig. \ref{fig:lidar_fov}, PandarQT deserves a position on the very top to observe robot's near surrounding, including objects at very low angle, with no obstruction. Located straight under the PandarQT, the front Ruby covers far away objects. Both LiDARs, as well as all rest of them, are connected to separate cluster nodes through Ethernet cables. Each of these nodes handles the control and data acquisition of the respective LiDAR. Robosense LiDARs are synced using the LiDAR synchronization channel mentioned in Sec. \ref{subsection:sync}, while LiDARs of other models use PTP protocol to get time base from the syncboard relayed by the cluster nodes.

One floor lower, the first (front) Robosense Bpearl is installed on the front panel. Its zenith of the field of view (FoV) points forward, providing a hemispherical view of the front of the robot, also shown in Fig. \ref{fig:lidar_fov}. 

\begin{figure}[htbp]
    \centerline{\includegraphics[width=\linewidth]{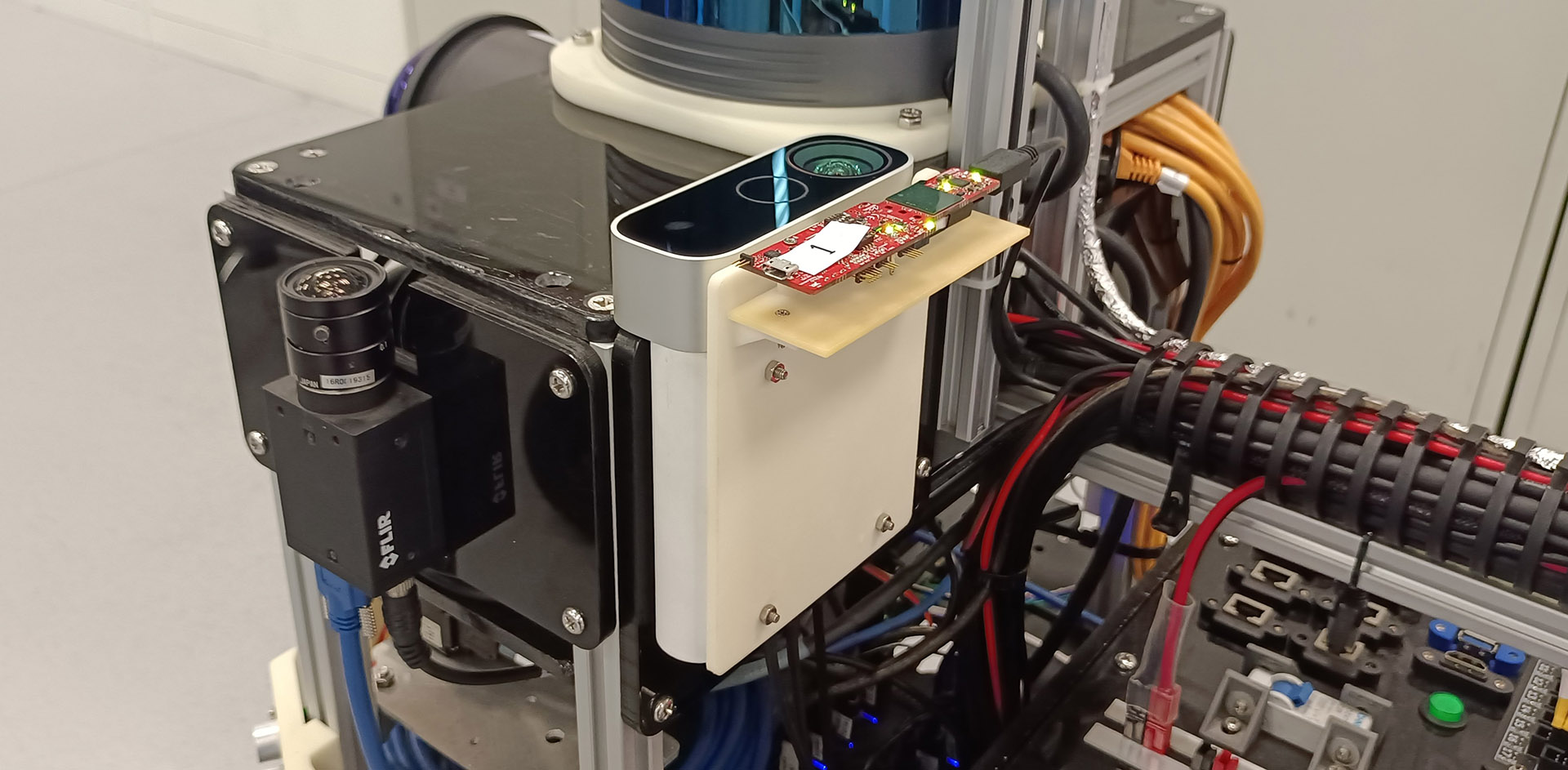}}
    \caption{The upward looking Azure Kinect DK and TI IWR6843AOPEVM mmWave radar.}
    \label{fig:kinect_radar_up}
\end{figure}

Behind the front Robosense Bpearl, 2 Grasshopper3 cameras, one Azure Kinect DK and a mmWave Radar, shown in Fig. \ref{fig:front_sensor_tower} and Fig. \ref{fig:kinect_radar_up} are attached on the two side panels and the back panel of this floor. They look upwards and provide a view of ceilings or treetop above. Inside this floor, there stashes the cables and I/O boxes of all LiDARs on the robot.

The two up-looking Grasshopper3 cameras, as well as their front-, left-, right- and back-looking counterparts, are equipped with Kowa LM6JC wide-angle lenses. They have a focal length of 6mm, providing a wide FoV of $81.9^{\circ} \times 61.2^{\circ}$ and a deep depth of view \cite{LM6JC}. We set the focal distance to $\infty$ and the aperture to their maximum value of $f/1.4$ to ensure a sharp imaging on far surrounding objects and let sufficient amount of light reach the sensors, enabling high frame-rate capturing.
These 10 cameras run in hardware triggering mode and are connected to the same triggering channel on the Syncboard, thus capture images at the same time, with same frame rate.

Yet another floor lower, there are two EvC3A event-based cameras, an MTi-630R IMU, Livox Avia LiDAR, and 13 KS103 ultrasound range finders. The two EvC3A event-based cameras form a stereo pair looking forward, using same wide FoV lenses as Grasshopper3's. The focal distance setting is the same but the aperture is set to the smallest since event-based sensors capture the change of brightness instead of the absolute brightness and work better with reasonably limited lighting. The two event-cameras synchronize their internal clock with each other using a 1 MHz signal and also receive pulse signals for time-stamping from the Syncboard. Instead of following the signals to capture whole frames as frame-based cameras would do, event-based cameras perceive them as 'External Trigger' events which would be stored in the same stream along with 'Contrast Detector' events mentioned in Sec. \ref{subsection:evc3a}.

The MTi-630R IMU is configured to measure linear acceleration, angular velocity, orientation and magnetic heading in all 3 axis, as well as atmospheric pressure and temperature, all at a rate of up to 400Hz triggered by the Syncboard.
The KS103 ultrasound range finders are installed on curved brackets attached to the tower to achieve evenly angled detecting directions. All 13 of them are connected via an I\textsuperscript{2}C bus to a microcontroller which triggers them in an interleaved sequence to avoid cross-talk.

At the bottom of front sensor tower, two A315 IR cameras and the Neuvition M1 LiDAR are mounted on the front end of the sensor deck. The A315 are equipped with 18mm lenses providing a 25° FoV and have motorized focus motors enabling automatic focusing control. We configured them to output a 16-bit linear temperature image at a frame rate of 60 Hz. They are connected over a Gigabit Ethernet connection to designated cluster nodes that handle their recording and controlling.
    
\begin{figure}[htbp]
    \centerline{\includegraphics[width=\linewidth]{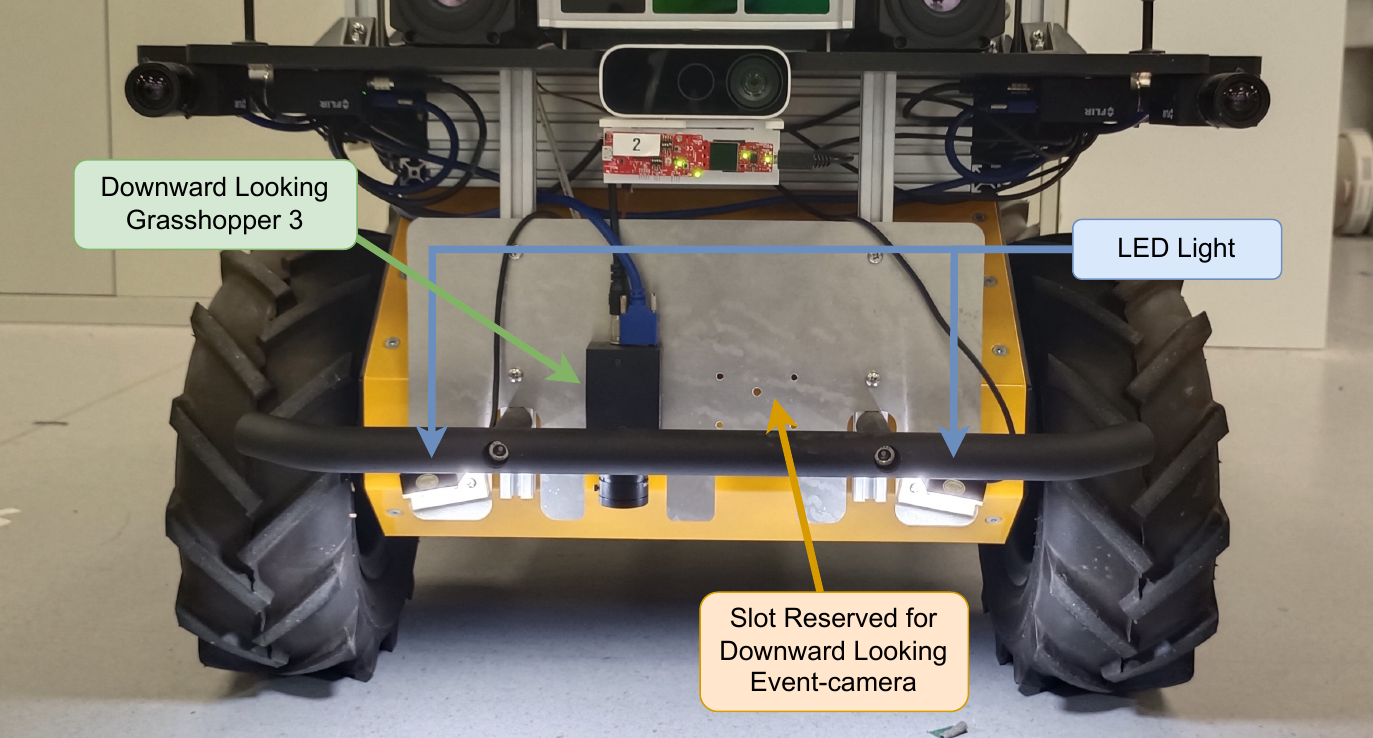}}
    \caption{The downward looking Grasshopper3 and LED lights.}
    \label{fig:downlooking_gs3}
\end{figure}

At the front of the UGV base, there is a downward looking Grasshopper3 camera, as shown in Fig. \ref{fig:downlooking_gs3}. Its focal distance is set to match the distance to the ground in order to get a clear view of the floor, road, etc. Alongside it, two 1.5-watt LED lights are installed to ensure enough lighting, enabling an even higher frame rate to compensate for the limited FoV (about $40\times 60$ cm on the ground) due to its low mounting position. We dedicated a triggering channel of the Syncboard for this camera to drive it at a different, often faster, frame rate than its counterparts looking around and upward.

\subsection{Sensor Deck}

\begin{figure}[htbp]
    \centerline{\includegraphics[width=\linewidth]{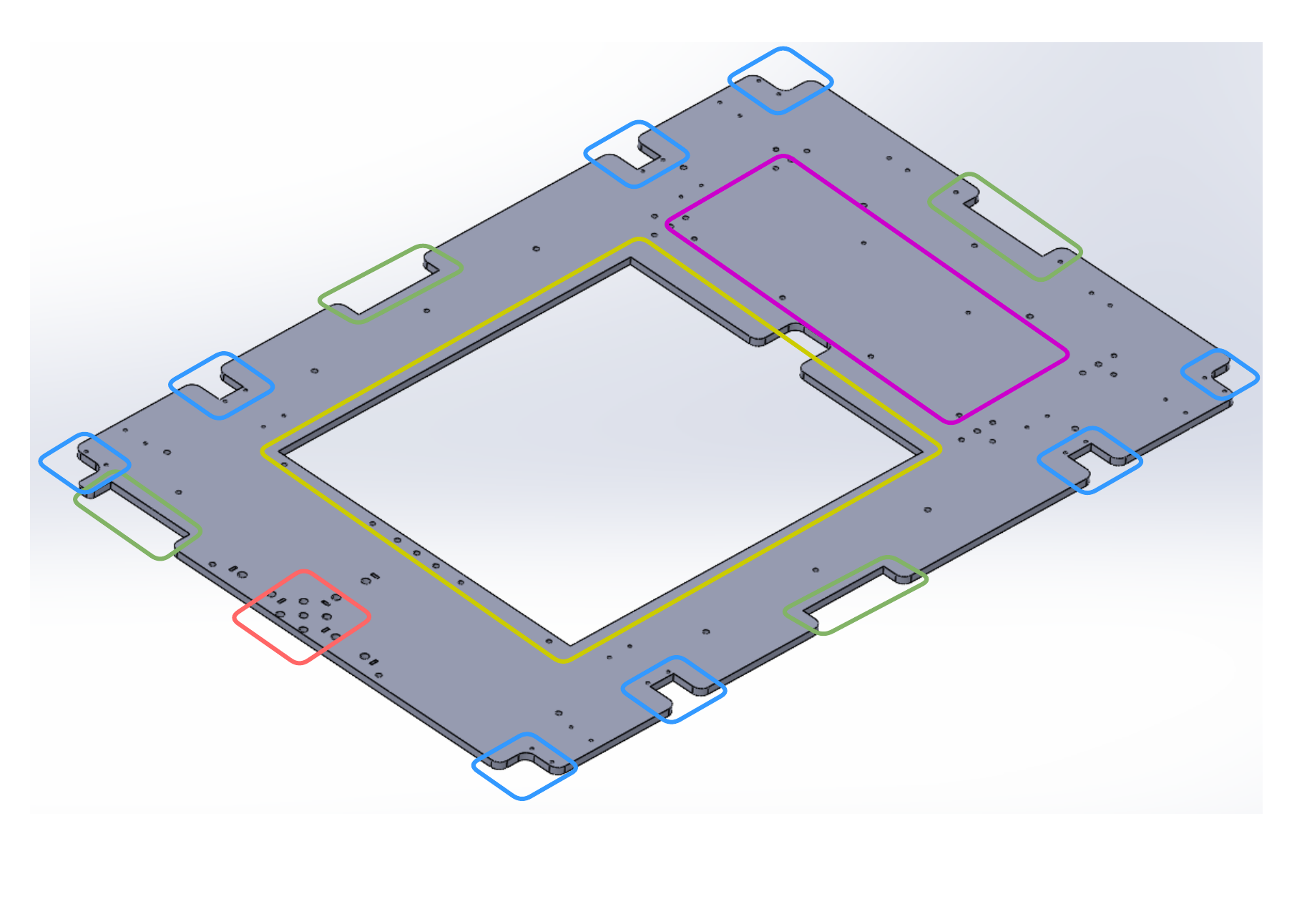}}
    \caption{The design of ShanghaiTech Mapping Robot's sensor deck. Mounting position for Grasshopper3 (blue), Azure Kinect DK (green), computation cluster (yellow), front sensor tower (purple) and back sensor tower (red).}
    \label{fig:sensor_deck}
\end{figure}

The 94cm x 63cm sensor deck is CNCed from aircraft grade 7050 aluminum of 10mm thickness and anodized to reduce reflections that may interfere with the sensor data. The deck's CAD is available alongside the dataset.
This deck defines the length and width of the vehicle.
It it mounted directly onto the base UGV by being bolted to the main aluminum extrusion frame around the top of UGV's payload bay by two 20mm x 80mm x 380mm aluminum profiles and further supported by aluminum pillars attached along with the front and rear bumper bars to the main chassis of the Husky UGV. Both the front and back sensor towers are attached to the sensor deck. The computation cluster, power box and one of the two batteries are placed through a 45cm x 38cm opening of the sensor deck, marked yellow in Fig. \ref{fig:sensor_deck}, into the UGV's payload bay.
            
The sensor deck accommodates the Grasshopper3 cameras, the Azure Kinect DK depth cameras and the mmWave radars facing the front, back, left and right directions.
In each direction, the two separate Grasshopper3 mounting slots, marked blue in Fig. \ref{fig:sensor_deck}, are capable of aligning the direction and sensor plane of cameras, forming them into a stereo pair.

\begin{figure}[htbp]
    \centerline{\includegraphics[width=\linewidth]{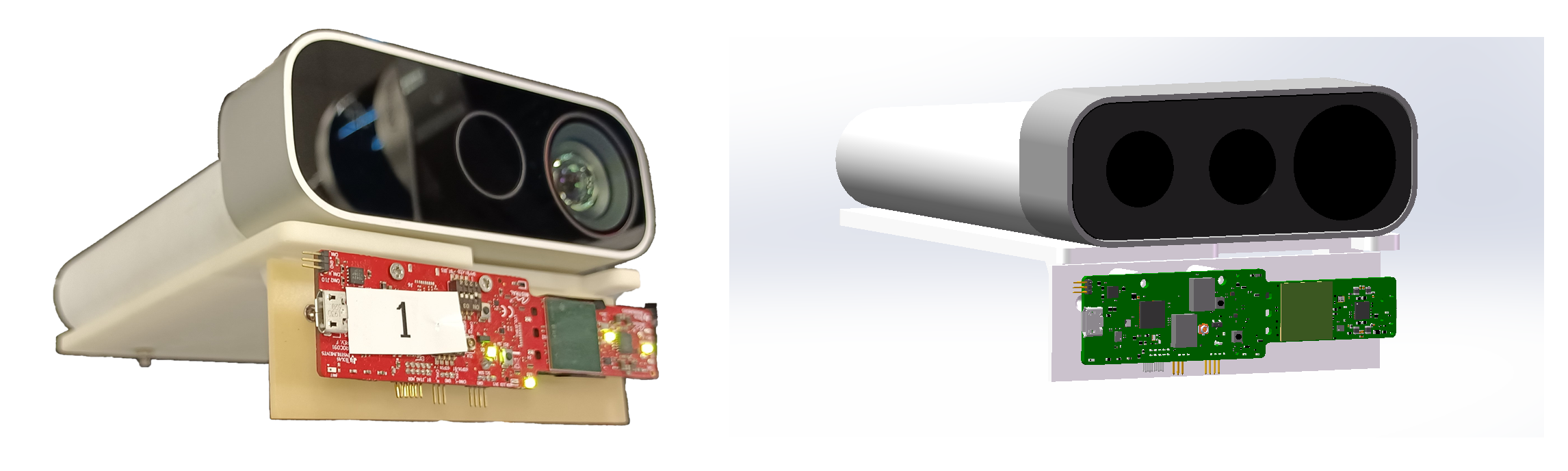}}
    \caption{The design (right) and actual photo (left) of 3D printed part connecting Azure Kinect DK and mmWave radar.}
    \label{fig:kinect_mmwave_pair}
\end{figure}

As shown in Fig. \ref{fig:kinect_mmwave_pair}, each mmWave radar is mounted on the back of a Azure Kinect DK using a 3D printed part, forming a Kinect-mmWave pair. Such pairs are attached to Kinect mounting slots of sensor deck, marked green in Fig. \ref{fig:sensor_deck}.
Such approach ensures a uniform translation between all 5 pairs and thus helps simplifying the calibration of mmWave radar extrinsics. While mmWave radars do not support hardware triggering, Azure Kinect DKs do. They are all configured to run in the subordinate mode and have their sync-in ports connected to a dedicated triggering channel of the Syncboard generating the Kinect's VSync signal at desired frame rates.

The real-time kinematics GNSS receiver is placed on the sensor deck near the back sensor tower. Its two antennas are installed atop the front and back sensor tower, respectively. The receiver is connected to the cluster LAN and thus has access to online differential corrections data via a 5G cellular link from either local base station or the third-party CORS (Continuously Operating Reference Stations) network. It also sends a PPS signal to the Syncboard via a dedicated cable, such that it can  act as an external master clock,  if we configure it this way, as mentioned in Sec. \ref{subsection:sync}.

The powering, data and synchronization cables of the deck's sensors are routed in cable channels on the bottom sides of the deck.

\subsection{Back Sensor Tower}

The back sensor tower is made out of a single $40 \times 40 \times 800$ mm aluminum profile pillar.
Along the pillar, two rotational LiDARs, the omni-directional camera, the rear GNSS antenna as well as WiFi, cellular and joystick receivers are attached. 

\begin{figure}[htbp]
    \centerline{\includegraphics[width=\linewidth]{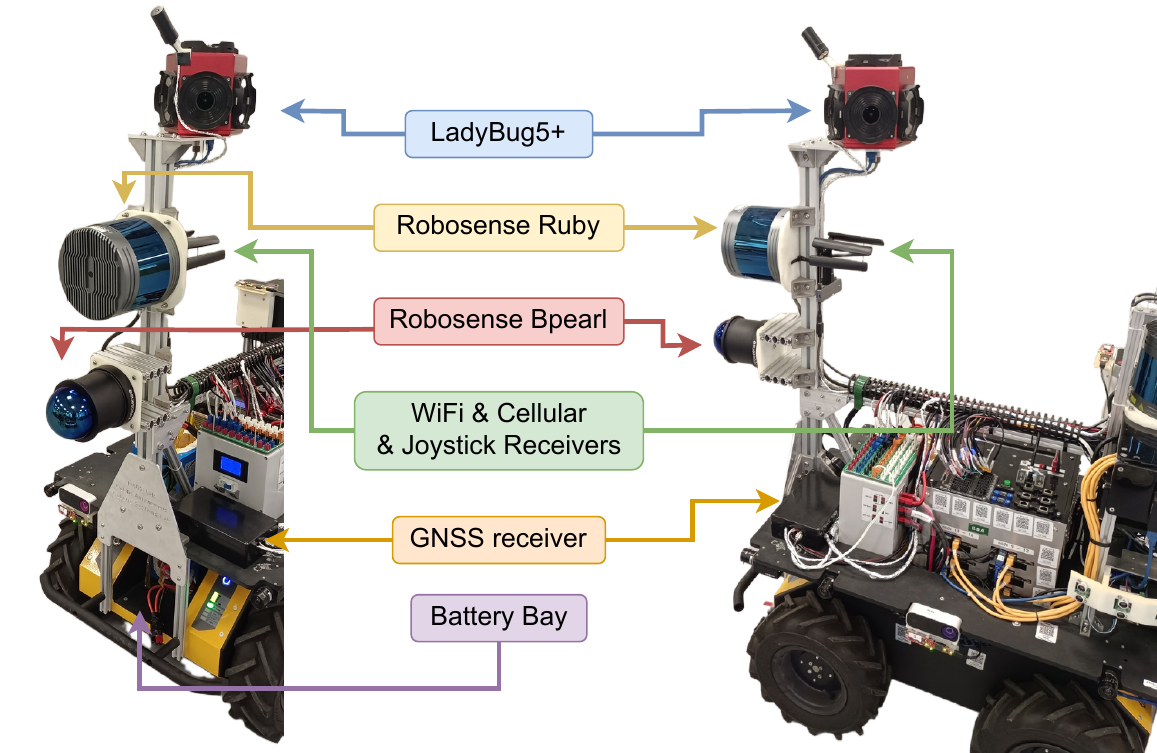}}
    \caption{The back sensor tower of ShanghaiTech Mapping Robot viewed from its back (left) and right (right).}
    \label{fig:back_sensor_tower}
\end{figure}

The omni-directional LadyBug5+ is mounted at the top of the back sensor tower.
The built-in lenses of LadyBug5+ have a $f/2.5$ aperture and a 4.4mm focal length, achieving a FoV of 90\% full sphere (6 sensors combined) and acceptable sharpness from about 60cm to infinity \cite{ladybug5p}. We capture original pinhole images from the 6 sensors rather than using its SDK to stitch panorama images on the fly, for preserving the raw data and to save computation time. It is triggered via a dedicated channel of Syncboard. The back GNSS antenna is placed on the top of LadyBug5+ in a particular orientation that hides it in a blind spot.

\begin{figure}[htbp]
    \centerline{\includegraphics[width=\linewidth]{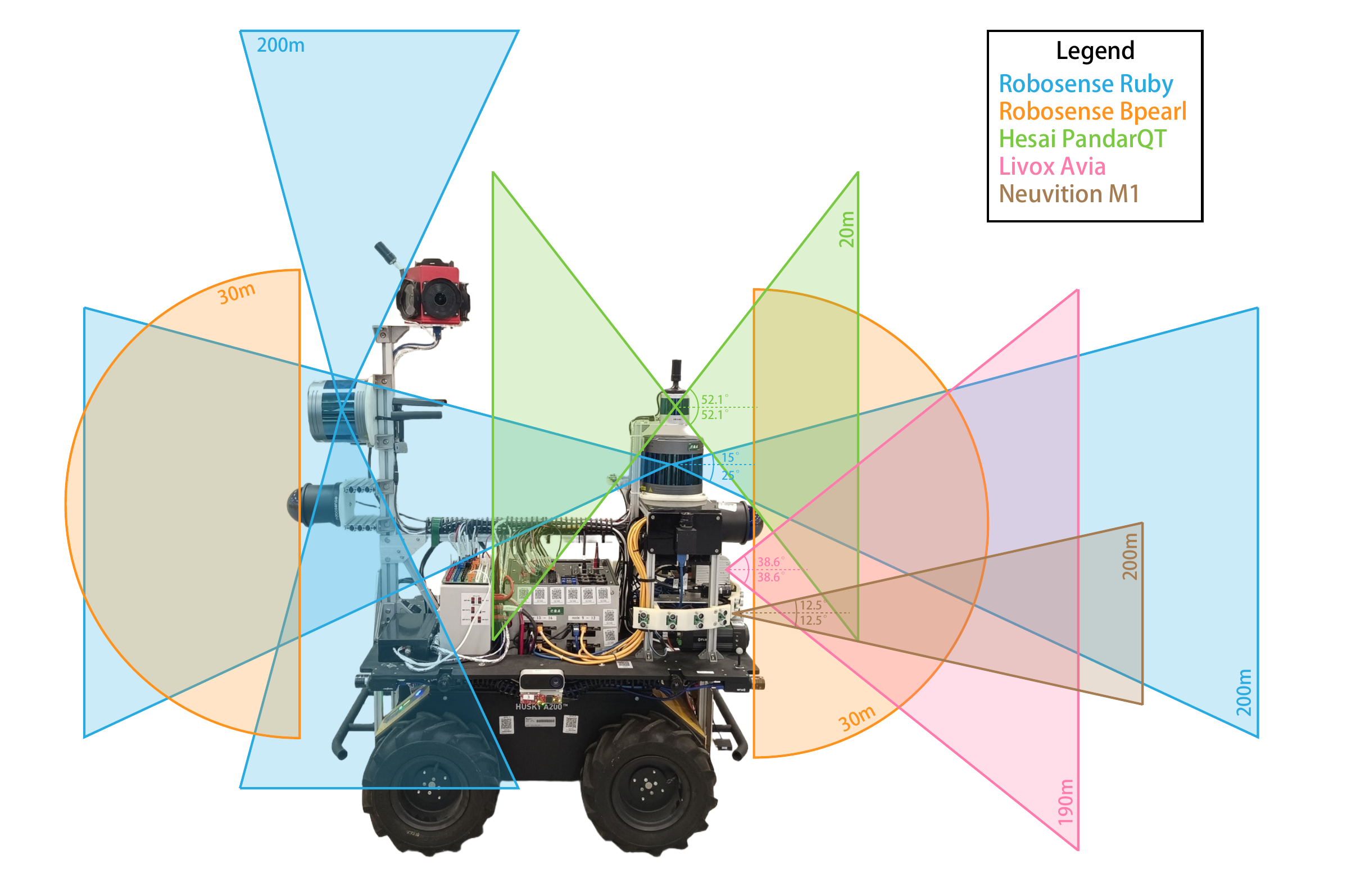}}
    \caption{FoV of LiDARs equipped by ShanghaiTech Mapping Robot. }
    \label{fig:lidar_fov}
\end{figure}

The second back Robosense Ruby and Bpearl are mounted in the middle of the tower. The back Bpearl simply points the zenith of its FoV backwards, opposite to its front counterpart, while the back Ruby is installed vertically, pitched up 90°, achieving a dense vertical scanning and coverage of far objects with high elevation angle like tall buildings along the routes. In Fig. \ref{fig:lidar_fov}, we show the FoV of all 7 LiDARs on the robot.

\begin{figure}[htbp]
    \centerline{\includegraphics[width=\linewidth]{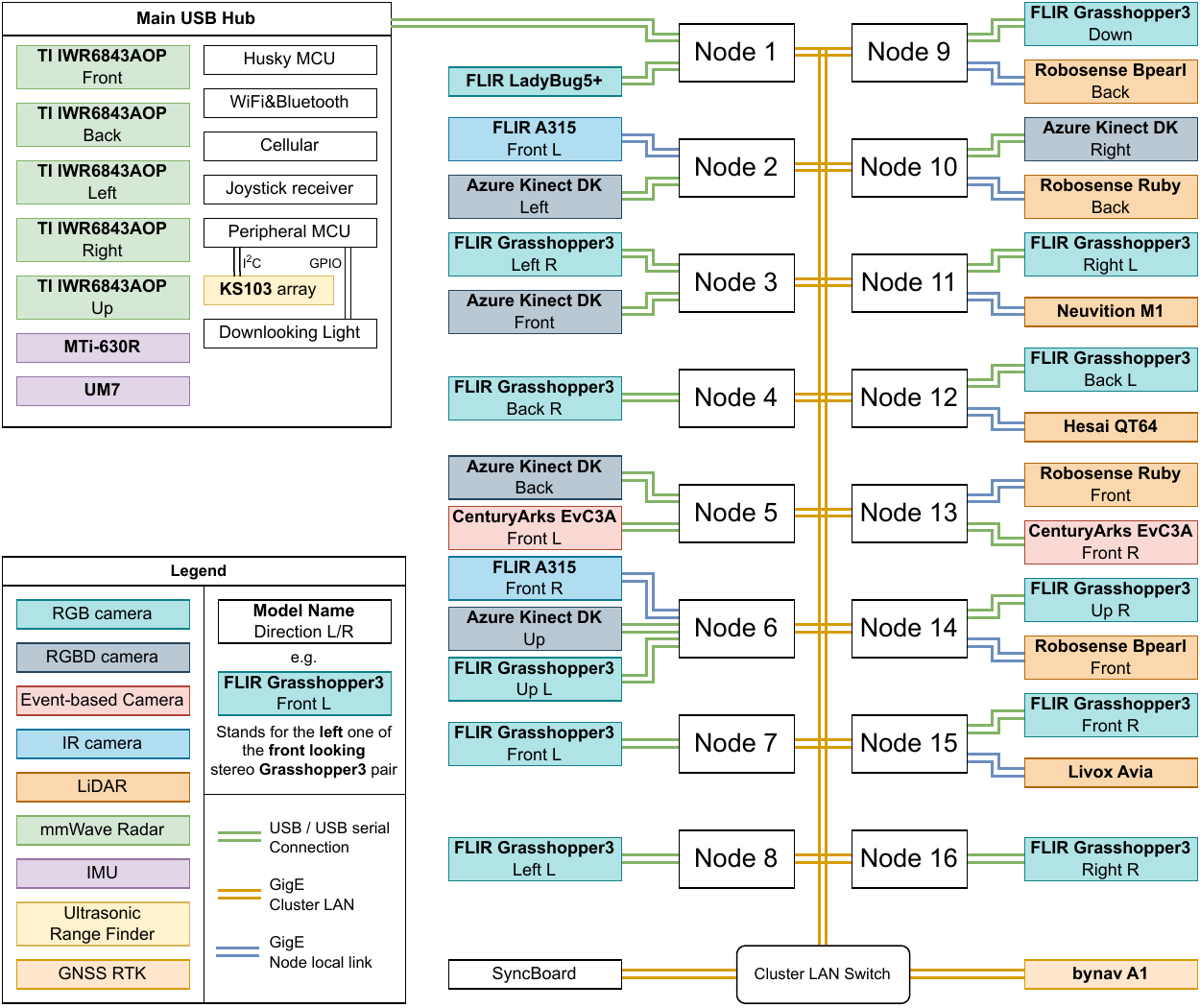}}
    \caption{The communication structure of the ShanghaiTech Mapping Robot. }
    \label{fig:comm_struct}
\end{figure}

\begin{figure}[htbp]
    \centerline{\includegraphics[width=\linewidth]{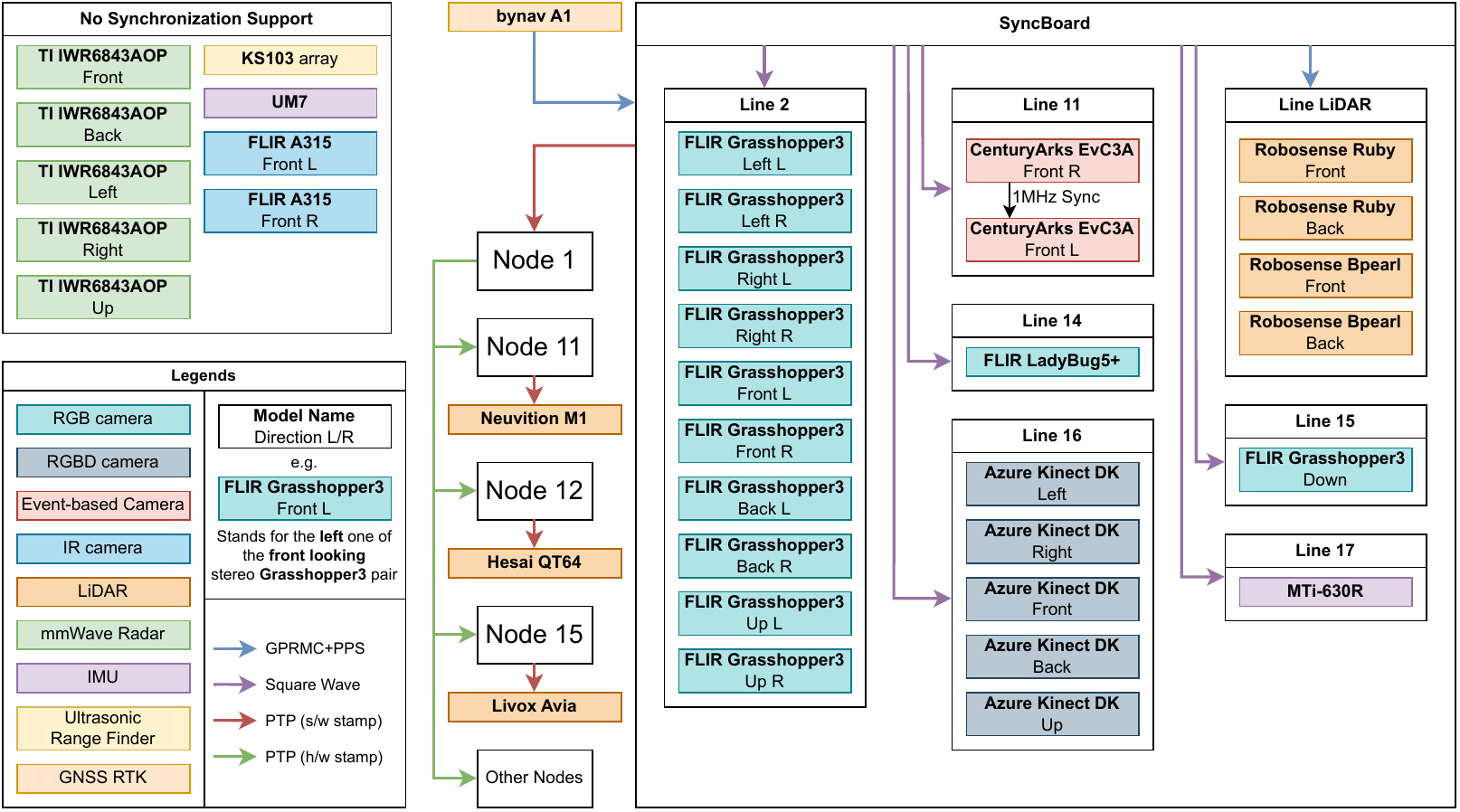}}
    \caption{The synchronization flow of the ShanghaiTech Mapping Robot. }
    \label{fig:sync_flow}
\end{figure}

A USB hub is installed on the back sensor tower to accommodate WiFi, cellular and joystick receivers, providing connectivity to the robot for controlling, monitoring, and access to online reference servers for time and differential GNSS.
Below the back sensor tower, a second battery is installed in the back bay of UGV.
With everything connected, the communication and synchronization architectures of the whole robot are as shown in Fig. \ref{fig:comm_struct} and Fig. \ref{fig:sync_flow}.

\section{Calibration}
\label{sec:calibration}

To ensure accurate perception and mapping, it is crucial to calibrate the intrinsic and extrinsic parameters of the various sensors employed in the system. There are various tools suitable for our use case, including ones that concentrate on sensors of one same type \cite{6696514}, \cite{tahiraj2024gmmcalib}, \cite{RAMEAU2022103353}, and others performing cross-modal calibration \cite{10160691}, \cite{kalibrCamIMU}, \cite{chy230}, \cite{wang2022accurate}, \cite{zhi2022multical} and \cite{zhu2022robust}. This section outlines the calibration procedures for the RGB cameras, event-based cameras, and LiDAR sensors.

\subsection{Camera Intrinsic \& Extrinsic Calibration}

The intrinsic and extrinsic parameters of 21 out of the 22 RGB cameras, including 10 Grasshopper3, 5 Azure Kinect DK, and 6 individual sensors of the LadyBug5+, but excluding the downlooking Grasshopper3, were calibrated using the multi-camera calibration toolbox Kalibr \cite{6696514} and AprilGrid patterns. Kalibr is capable of calibrating the intrinsic and extrinsic parameters of camera systems with non-globally shared overlapping fields of view and supports various distortion models, such as radial-tangential and equidistant models, which accurately represent the lenses used on the Grasshopper3 and the built-in fisheye lenses on the Ladybug5+.

For each of the 5 main directions (front, back, left, right, and up), the two Grasshopper3s, one Azure Kinect DK, and one or two sensors of the LadyBug5+ facing that side are grouped and calibrated together since they share a similar perspective. This gives us intrinsics of all cameras and 5 extrinsics groups, each independent from the others. Subsequently, the extrinsics between the 5 surrounding LadyBug5+ sensors were calibrated in a separate run, connecting the extrinsics of the 4 horizontal sides (front, back, left, and right). Because all up-looking and down-looking cameras do not share sufficient overlap with any side-looking cameras, Kalibr cannot be used to estimate their extrinsics. Thus the up-looking side was finally connected to the extrinsic tree using the factory pre-calibrated transformation between the front-looking and up-looking sensors of the LadyBug5+.

\begin{figure}[htbp]
    \centerline{\includegraphics[width=\linewidth]{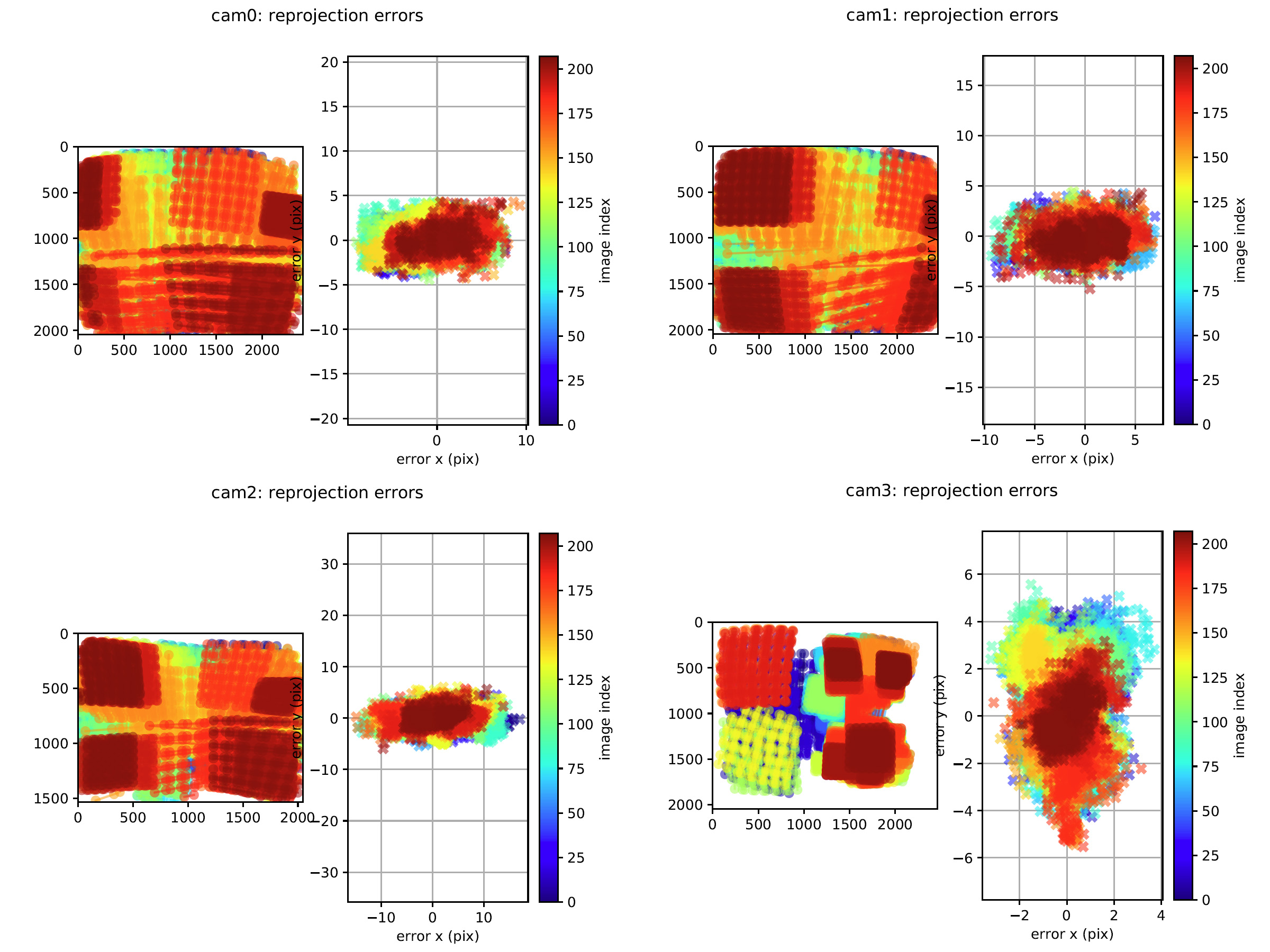}}
	\caption{Reprojection error of 4 frontward looking RGB cameras after calibration by Kalibr. (cam0: stereo left Grasshopper3, cam1: stereo right Grasshopper3, cam2: Azure Kinect DK RGB sensor, cam3: LadyBug5+ sensor number 0). }
	\label{fig:reprojection_error_front}
\end{figure}

The intrinsic parameters of the last, down-looking Grasshopper3 camera were calibrated using the ROS camera calibration package \cite{roscalib} and a checkerboard target with a size matching the small area of view on the ground. Considering that the camera has no overlapping field of view with any other onboard sensors and that it will not be used for 3D SLAM but rather down-looking odometry, its extrinsic parameters were derived from the structure design instead of using complex calibration algorithms.

The intrinsic parameters of the two EvC3A event-based cameras were calibrated using blinking checkerboard patterns and the calibration tool provided by the Metavision SDK. The extrinsics of the event-cameras were calibrated using Kalibr, utilizing the frame images rendered from the two event streams along with the two image streams captured by the two front-facing Grasshopper3 cameras.

\subsection{LiDAR Extrinsic Calibration}

The extrinsics between the LiDARs and cameras were calibrated using a combination of LiDAR-Camera and LiDAR-LiDAR approaches. The extrinsic between the front-looking camera of the Ladybug5+ and the Livox Avia LiDAR was estimated using the direct visual-LiDAR calibration toolbox \cite{10160691}. The Livox Avia's non-repetitive scanning pattern provides a dense point cloud of the scene in front of the stationary robot, which the toolbox utilizes for accurate direct pixel-level registration with the wide-angle camera image captured by the LadyBug5+.

To calibrate the extrinsics between LiDARs, we registered the scans of each LiDAR against a high-precision scan from a FARO Focus 3D scanner \cite{faro}. The LiDAR scans were captured with the robot standing still and then averaged to reduce temporal noise. The registration algorithm, Iterative Closest Point (ICP) to be specific, yields the transformations between LiDARs. Combined with the camera-LiDAR transformation, all LiDARs were connected to the extrinsics tree.

\subsection{Calibration of Other Sensors}

The MTi-630R IMU's Allan Variance parameters, including random walk and bias instability, were calibrated using the allan\_variance\_ros tool \cite{allanVariance}. With the IMU intrinsics, the Kalibr toolbox calibrated the extrinsic between the IMU and the two front-looking Grasshopper3 cameras. The mmWave radars on each side share an identical extrinsic with the respective Azure Kinect camera they are attached onto, derived from the 3D printed mount's design. The extrinsics of the ultrasound range finders and other unmentioned sensors were either roughly measured manually or derived from their mounts' structure design.

\begin{figure}[htbp]
    \centerline{\includegraphics[width=\linewidth]{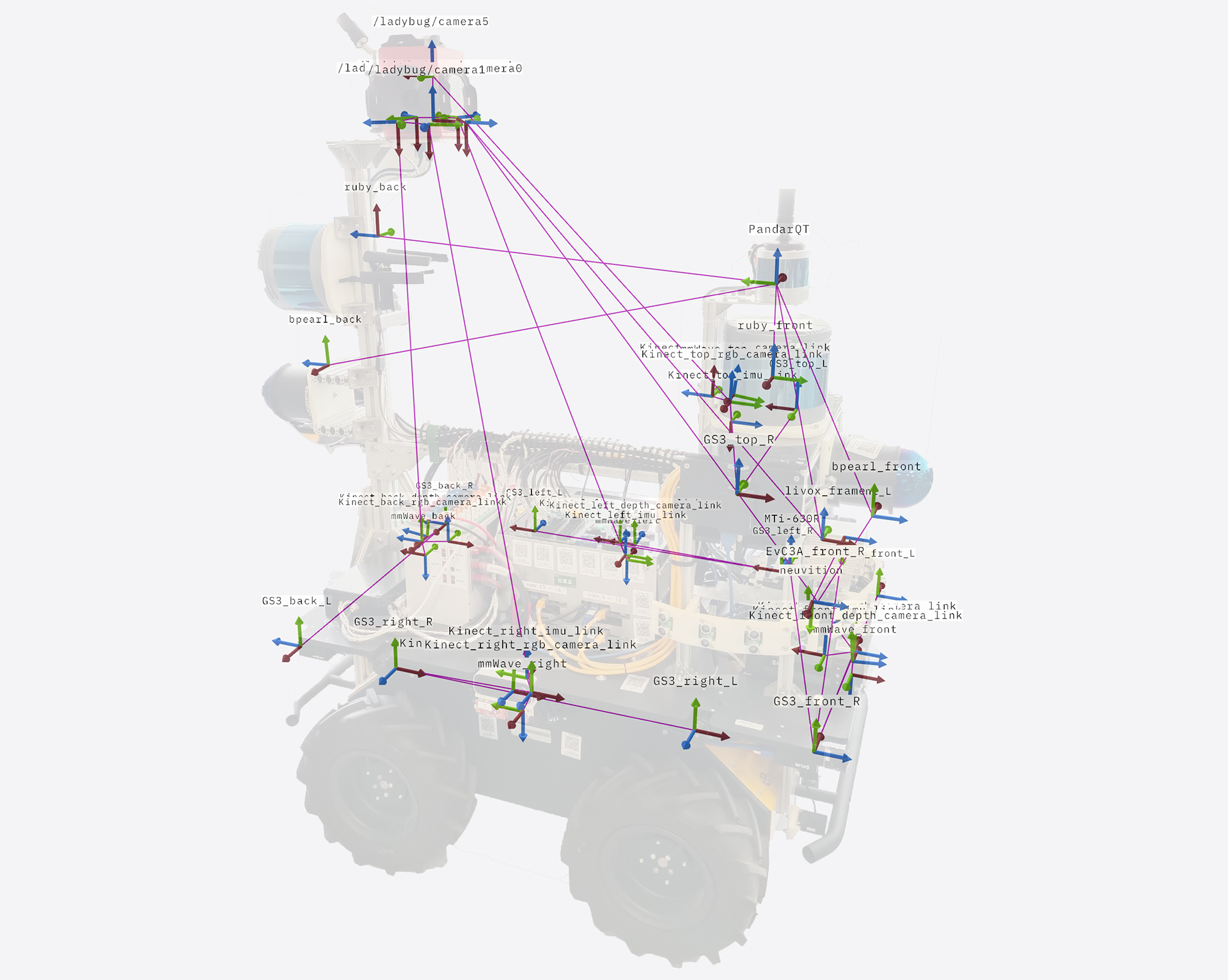}}
	\caption{Coordinates of calibrated sensors. Purple lines indicate the edges of transformation tree. An image of mapping robot taken from a similar perspective is also shown. }
	\label{fig:calibration_extinsics}
\end{figure}

\section{Procedures and Post-Processing}
\label{sec:procedures}

The mapping robot is designed with distributed computation in mind, so it is vital to orchestrate all the nodes in the computation cluster along with the sensors connected to them. In this section we introduce the procedure of data acquisition and post-processing to be performed on the collected data.

\subsection{Procedure of Data Acquisition}

\subsubsection{Material Preparation}

At minimum, the robot itself, one battery, and a controlling terminal like KVM (Keyboard VGA Mouse) or laptop is needed for performing a acquisition run. A second battery as well as backup ones or power adaptor is recommended.

\subsubsection{Power Up}

Closing the main breaker on the power box, the Husky UGV, the nodes of the cluster, the Syncboard as well as sensors will be provided with electricity from the battery, as mentioned in Sec. \ref{sec:powerbox}. Only the master node (Node 1) and Syncboard boot when power becomes available. After Node 1 boot process is finished, the robot becomes maneuverable.

\subsubsection{Cluster Bringup}

The master node sets itself up as the Ethernet gateway and network file systems for Boot-from-LAN, then wakes up all the worker nodes. Once they have booted up, we can control the nodes and access data stored in them. Network access also becomes available for nodes, the Syncboard and the GNSS receiver.

\subsubsection{Time Synchronization}

The Syncboard gets the time base from an external master clock and broadcasts it over the cluster LAN. We use scripts to trigger immediate time synchronization on the nodes and print their clock times for manual verification, then launch the triggering for sensor testing procedure of the next step.

\subsubsection{Sensors Bringup}

The master node orchestrates all worker nodes to launch the sensor drivers. We would manually check if all sensors work as intended. If that's not the case, sensor's parameters would be modified accordingly. After everything is confirmed to work well, the triggering is turned off awaiting the upcoming data acquisition.

\subsubsection{Data Acquisition}

Here we command the master node to orchestrate a distributed recording of sensor data. Then we turn on triggering to officially start a data acquisition run. We would now drive the robot along the data acquisition route.

\subsubsection{End of Acquisition}

After the robot reaches the finishing position of the route, the triggering is turned off first, followed by the termination of the data recording processes. Sensor drivers stay running and do not have to be spawned repetitively until there is no more data acquisition runs to be carried out before powering off the robot. The data collected can be copied either into external hard drives via a USB port from each node or to a workstation or network storage device through Ethernet connection, pending for post-processing.
    
\subsection{Dataset Post-processing}

\subsubsection{Timestamp Restoration}

In order to be able to utilize the actual trigger times of the sensor data, which is important for many algorithms like stereo calculation or sensor fusion, especially with the IMUs working at a high frequency, we re-timestamp all those messages with the time of the actual trigger, as recorded by the Syncboard, rather than the time the data was processed on the node. This is done for the messages from all hardware-triggered sensors (Grasshopper3, LadyBug5+, Kinect Azure and MTi-630R IMU) shown in Fig. \ref{fig:sync_flow}. Any frame drop would be discovered and noted at the same time.

\subsubsection{Bayer Interleaving}
LadyBug5+ on-the-fly compresses each image for the 6 camera sensors in 4 separate Bayer channels using JPEG, resulting in 24 JPEG images per one capture. We decompress and interleave them to get 6 image streams, one for each sensor, in bayer\_rggb8 format.

\subsubsection{Image Compression}
To provide an alternative compressed version for faster distribution of our dataset, we debayer each image stream into YUV format and compress them with our ROS to AV1 encoding tool \cite{videocompression}.

\subsubsection{Mapping Ground Truth}
As an optional step, for acquiring 3D ground truth point clouds, especially also for estimating the indoor ground truth robot trajectory, a FARO Focus 3D scanner can be used to obtain a precise point cloud of the environment along the data acquisition route, acting as the mapping or 3D reconstruction ground truth.

\subsubsection{Trajectory Ground Truth}
For route sections where GNSS signal is available, we will post-process the onboard differential GNSS/INS data to obtain a ground truth trajectory with a typical precision of 1cm. Other than that, we register LiDAR scans to the mapping ground truth to get the pose of robot as an alternative ground truth for indoor datasets.

\section{LiDAR \& RGB-D Interference Evaluation}
\label{sec:Interference}

\begin{figure}[htbp]
    \centerline{\includegraphics[width=\linewidth]{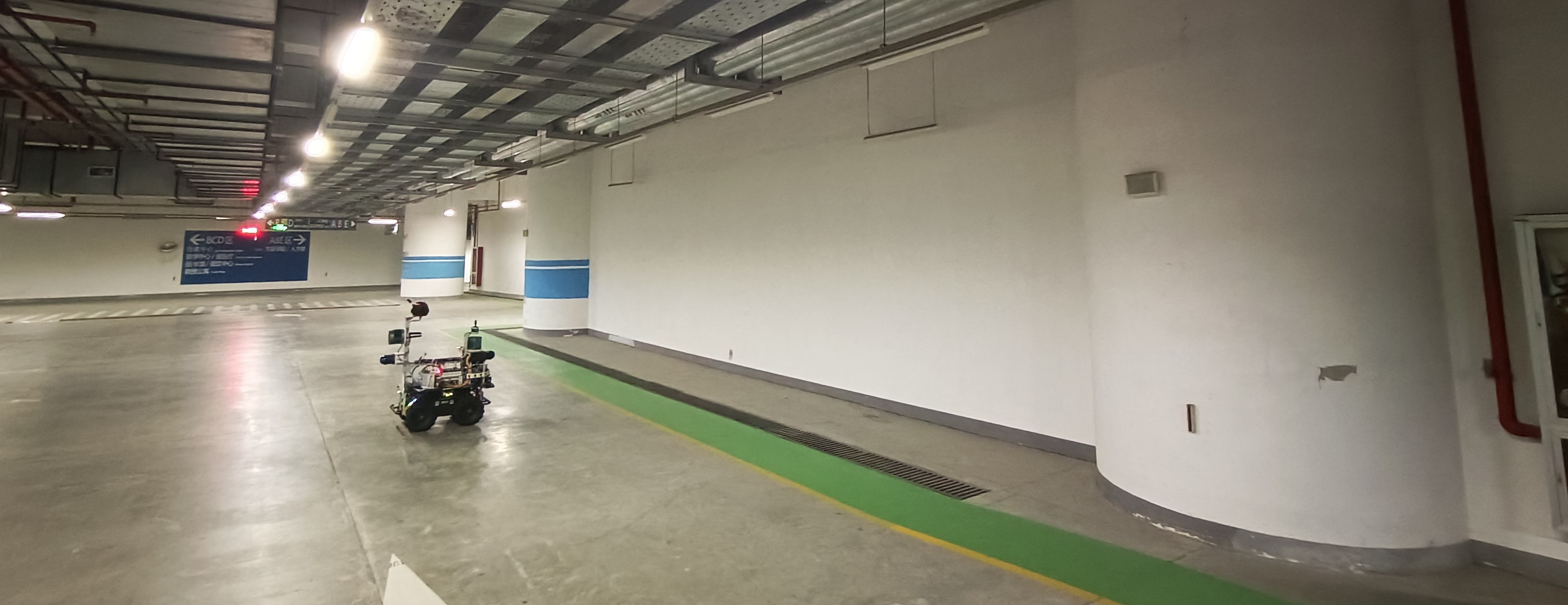}}
	\caption{Interference experiment scene setup.}
	\label{fig:interference_scene}
\end{figure}

LiDARs and RGB-Depth cameras on the mapping robot emit infra-red beams and measure the time of flight to obtain distance readings. There are concerns about potential interference between the sensors. In this experiment, we collected point clouds and depth images at a fixed position against a flat, white wall in the basement of ShanghaiTech University with each LiDAR and two RGB-Depth cameras working alone (sole runs) and together (collective run) for 10 seconds, as shown in Fig. \ref{fig:interference_scene}.

\subsection{Precision Analysis}

For rotational LiDARs, the temporal mean and standard deviation of distance measurements for each beam are calculated to indicate the average measurement and precision in respective directions. The mean and standard deviation of the temporal standard deviation among all beams are then computed to indicate overall precision. For unconventional LiDARs (Livox Avia and Neuvition M1), the FoV is binned in elevation and azimuth directions, and the nearest reading in each bin of each ROS pointcloud message is used. For the Azure Kinect depth camera, statistics are performed on each pixel.

\begin{equation}
	\begin{aligned}
		\overline{D_{b}} &= \frac{1}{N_t}\sum_{t=1}^{N_t}D_{t,b} \\
		\sigma_b &= \sqrt{\frac{1}{N_t-1}\sum_{t=1}^{N_t} (D_{t,b} - \overline{D_{b}})^2} \\
		\overline{\sigma_b} &= \frac{1}{N_b}\sum_{b=1}^{N_b}\sigma_b \\
		\sigma(\sigma_b) &= \sqrt{\frac{1}{N_b-1}\sum_{t=1}^{N_b} (\sigma_b - \overline{\sigma_b})^2}
		\label{eq:interference}
	\end{aligned}
\end{equation}

Equation \ref{eq:interference} is used for calculating the figures for LiDAR precision analysis, where $D_{t,b}$ is the distance measurement of beam or bin $b$ at time $t$, and $\overline{\sigma_b}$ is the mean of its temporal standard deviation.

\begin{table*}[htbp]
	\renewcommand{\arraystretch}{1.3}
	\caption{Interference Evaluation: Mean \& S.D. of temporal S.D. of distance\\
		Bold figures indicates significant interference}
	\label{table:interference_precision}
	\centering
	\begin{tabular}{c|c|c|c|c}
		\hline
		 Sensor &  Sole $\overline{\sigma_b}$ (m) &  Collective $\overline{\sigma_b}$ (m) &  Sole $\sigma(\sigma_b)$ (m) &  Collective $\sigma(\sigma_b)$ (m)\\
		\hline
		Robosense Ruby (front) & 0.0165 & 0.0156 & 0.1961 & 0.3605 \\
		Robosense Ruby (back) & 0.0104 & 0.0104 & 0.3713 & 0.3420 \\
		Robosense Bpearl (front) & 0.0114 & \textbf{0.1719} & 0.1225 & \textbf{1.4642} \\
		Robosense Bpearl (back) & 0.0278 & \textbf{0.0690} & 0.1970 & \textbf{0.8071} \\
		Hesai PandarQT & 0.0367 & 0.0373 & 0.2507 & 0.2507 \\
		Livox Avia & 0.0156 & \textbf{2.5236} & 0.0162 & \textbf{13.9601} \\
		Neuvition M1 & 0.0027 & 0.0028 & 0.0025 & 0.0026 \\
		Azure Kinect DK (front) & 0.0775 & 0.1357 & 0.2511 & 0.2693 \\
		Azure Kinect DK (up)& 0.2933 & 0.3064 & 0.5330 & 0.5361\\
		\hline
	\end{tabular}
\end{table*}

Table \ref{table:interference_precision} shows that all but the two Robosense Bpearls and the Livox Avia have consistent precision between the sole and collective runs, indicating that the precision of the other LiDARs is not affected by interference.

\begin{figure}[htbp]
	\subfigure[Side perspective]{\label{fig:interference_bpearl_a}
		\centerline{\includegraphics[width=\linewidth]{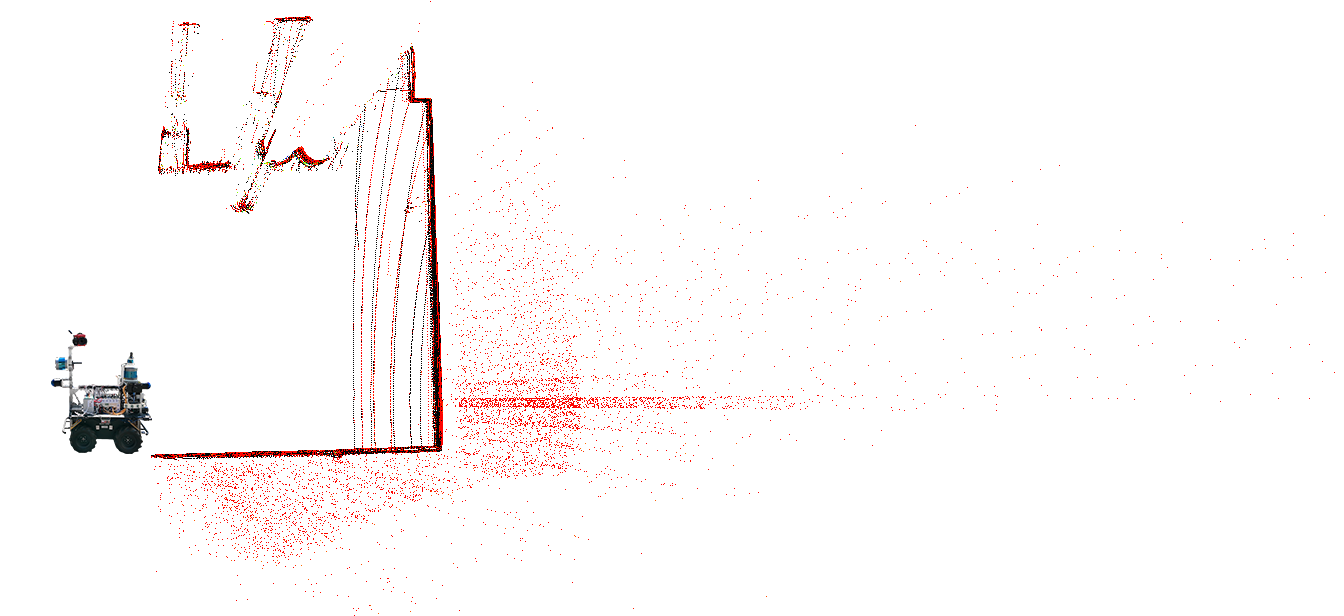}}
	}
	\subfigure[Top down perspective]{\label{fig:interference_bpearl_b}
		\centerline{\includegraphics[width=\linewidth]{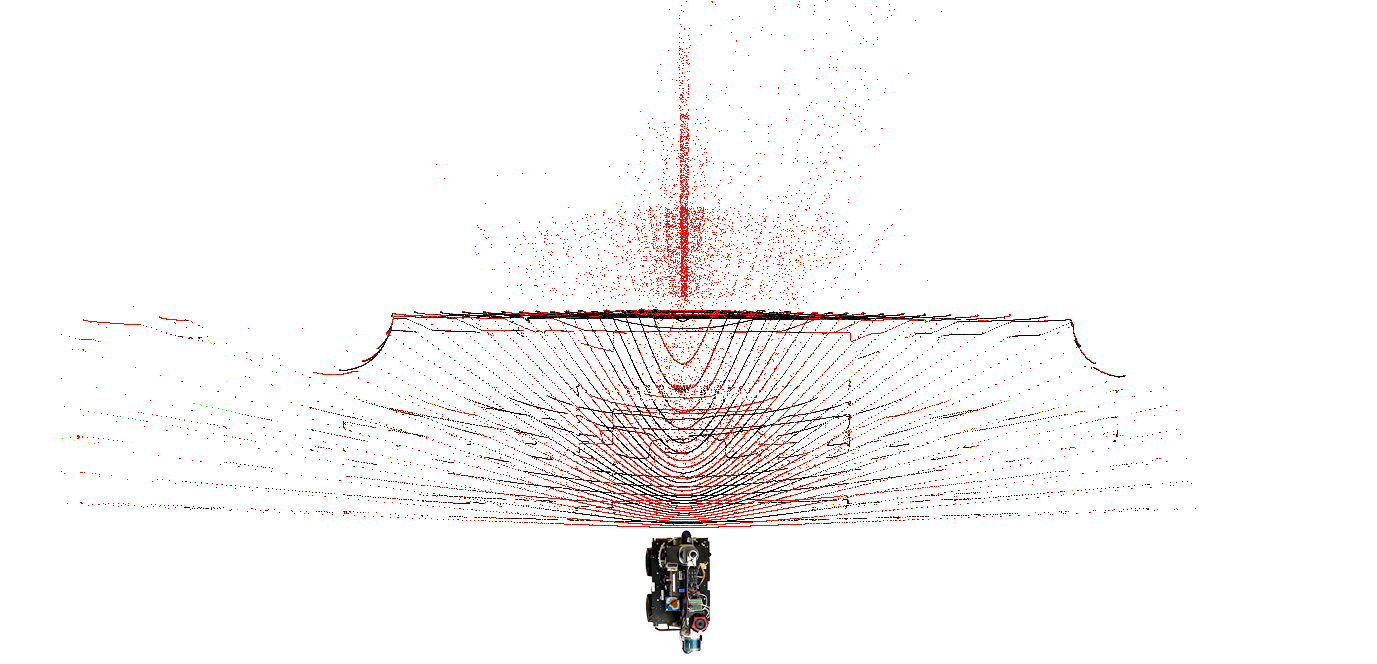}}
	}
	\caption{(a) 
 \& (b)
 Average point cloud produced by front Robosense Bpearl. Black points: sole run; Red points: collective run. The robot thumbnails indicate the approximate position of the robot and do not represent the actual scale.}
	\label{fig:interference_bpearl}
\end{figure}

Fig. \ref{fig:interference_bpearl}\subref{fig:interference_bpearl_a} \& \subref{fig:interference_bpearl_b} shows the temporal mean of the point cloud collected by the front Robosense Bpearl. In the collective run, both Robosense Bpearl and Livox Avia give noisy points behind the wall or below the floor, indicating interference. However, all valid objects are still detected by the LiDARs with substantial number of non-noisy points that have consistent standard deviation of distance compared to the sole run, suggesting that with proper filtering, the two sensors would give point clouds of the same precision regardless of interference.

\subsection{Accuracy analysis}

To assess the impact on accuracy, the distance between the sensors and a target plane is calculated. Significant changes in this distance would indicate interference-induced accuracy differences. The flat, white wall in front of the robot is used as the target plane for the 5 LiDARs on the front sensor tower, while another flat wall opposite to it is used for the back Robosense Bpearl, and the ground is used for the back Robosense Ruby.

For each sensor, point clouds from the sole and collective runs are merged separately. Points near the target plane are cropped out for accuracy evaluation. A reference plane is fitted to the selected points from the sole run, and the point-to-plane distance $D_{c2p}$ is calculated. The Gaussian distribution of this distance is then fitted for comparison.

\begin{table*}[htbp]
	\renewcommand{\arraystretch}{1.3}
	\caption{Interference Evaluation: Mean \& S.D. of Point-to-Plane distance}
	\label{table:interference_accuracy_lidar}
	\centering
	\begin{tabular}{c|c|c|c|c|c}
		\hline
		 Sensor &  Sole $\overline{D_{c2p}}$ (m) &  Collective $\overline{D_{c2p}}$ (m) &  Sole $\sigma(D_{c2p})$ (m) &  Collective $\sigma(D_{c2p})$ (m) & Rated Accuracy (m)\\
		\hline
  		Robosense Ruby (front, strongest return) & 0.0000 & -0.0357 & 0.0219 & 0.0239 & ±0.03\\
 		Robosense Ruby (front, last return) & 0.0000 & -0.0020 & 0.0143 & 0.0137 & ±0.03\\
		Robosense Ruby (back) & 0.0000 & -0.0004 & 0.0089 & 0.0088 & ±0.03\\
		Robosense Bpearl (front) & 0.0000 & 0.0022 & 0.0131 & 0.0130 & ±0.05\\
		Robosense Bpearl (back) & 0.0000 & 0.0009 & 0.0103 & 0.0078 & ±0.05\\
		Hesai PandarQT & 0.0000 & 0.0003 & 0.0111 & 0.0108 & ±0.03\\
		Livox Avia & 0.0000 & -0.0058 & 0.0115 & 0.0115 & ±0.02\\
		Neuvition M1 & 0.0000 & -0.0029 & 0.0140 & 0.0141 & ±0.02\\
		\hline
	\end{tabular}
\end{table*}

Table \ref{table:interference_accuracy_lidar} shows that the difference in point-to-plane distance between the sole and collective runs lies within $2\sigma$ for all LiDARs, indicating no significant accuracy change due to interference.

\begin{table}[htbp]
	\renewcommand{\arraystretch}{1.3}
	\caption{Interference Evaluation:\\ Pixel-wise Mean \& S.D. of Distance Difference}
	\label{table:interference_accuracy_kinect}
	\centering
	\begin{tabular}{c|c|c}
		\hline
		Sensor &  $\overline{D_{coll,p}-D_{sole,p}}$ (m) &  $\sigma(D_{coll,p}-D_{sole,p})$(m) \\
		\hline
		Kinect (front) & -0.0057 & 0.0332 \\
		Kinect (up)& -0.0001 & 0.0840 \\
		\hline
	\end{tabular}
\end{table}

For depth cameras, the difference in average distance measurements between the sole and collective runs is calculated pixel-wise to detect accuracy changes. As shown in Table \ref{table:interference_accuracy_kinect}, the mean difference is much smaller than its standard deviation, indicating that the depth cameras' accuracy is also unaffected by interference.

In conclusion, among all the infra-red light-emitting LiDARs and RGB-Depth cameras on the mapping robot, three give noisy outlier points that can be filtered out using statistical outlier removal approaches. Data from the other sensors along with the inlier data from the aforementioned three sensors have their accuracy and precision unaffected by interference. Except for the up-looking Kinect depth camera and the Hesai LiDAR, the sensors have no mutual sight, possibly preventing inescapable interference as the emitted infra-red laser does not directly shine into the photosensitive sensors of another without reflection from a third object.

\section{Outdoor Dataset}
\label{sec:outdoor}

To assess the mapping robot's data acquisition capability and verify the compatibility of the collected data with SLAM algorithms, we conducted an outdoor experiment on the ShanghaiTech University campus. The experiment involved collecting data along a predefined route and then running SLAM algorithms on the collected data.

\subsection{Data Acquisition}

\begin{figure}[htbp]
    \centerline{\includegraphics[width=\linewidth]{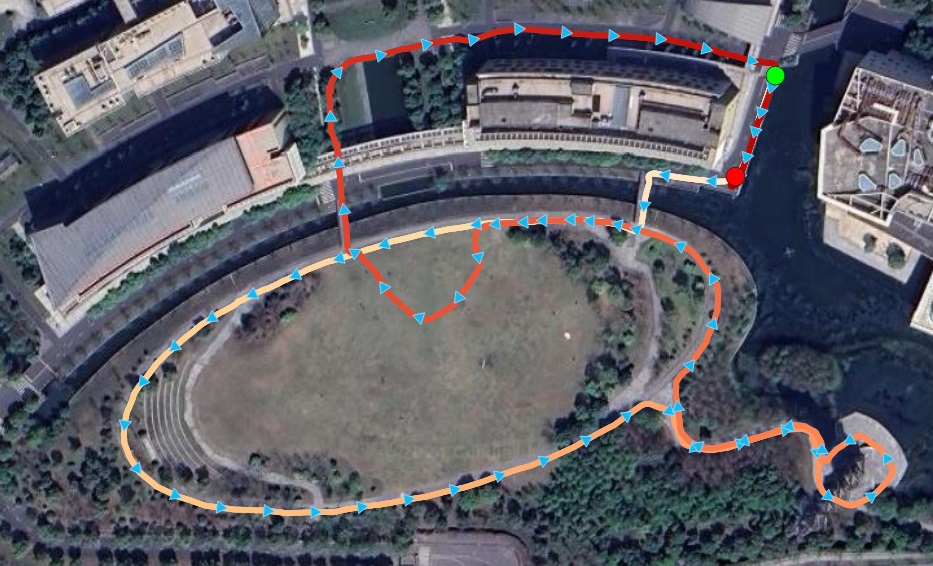}}
	\caption{Ground truth trajectory of the outdoor dataset, with its start (green) and end (red) point marked out. Color of the trajectory goes from light yellow to dark orange as time progresses.}
	\label{fig:outdoor_data_collection_gps_result}
\end{figure}

A 1.4 km data acquisition route, as shown in Fig. \ref{fig:outdoor_data_collection_gps_result}, was designed to include a variety of terrains and scenes on the ShanghaiTech University campus. The route was evenly split between areas around buildings and parkland, with the mapping robot primarily driving on paved roads or solid structures like bridges, and briefly on grass. Three sections of the route, surrounded by buildings, woods, or a blend of both, were revisited after traveling elsewhere, providing loop-closing opportunities for SLAM algorithms. The route also included slopes to enrich the elevation diversity.

\begin{figure}[htbp]
    \centerline{\includegraphics[width=\linewidth]{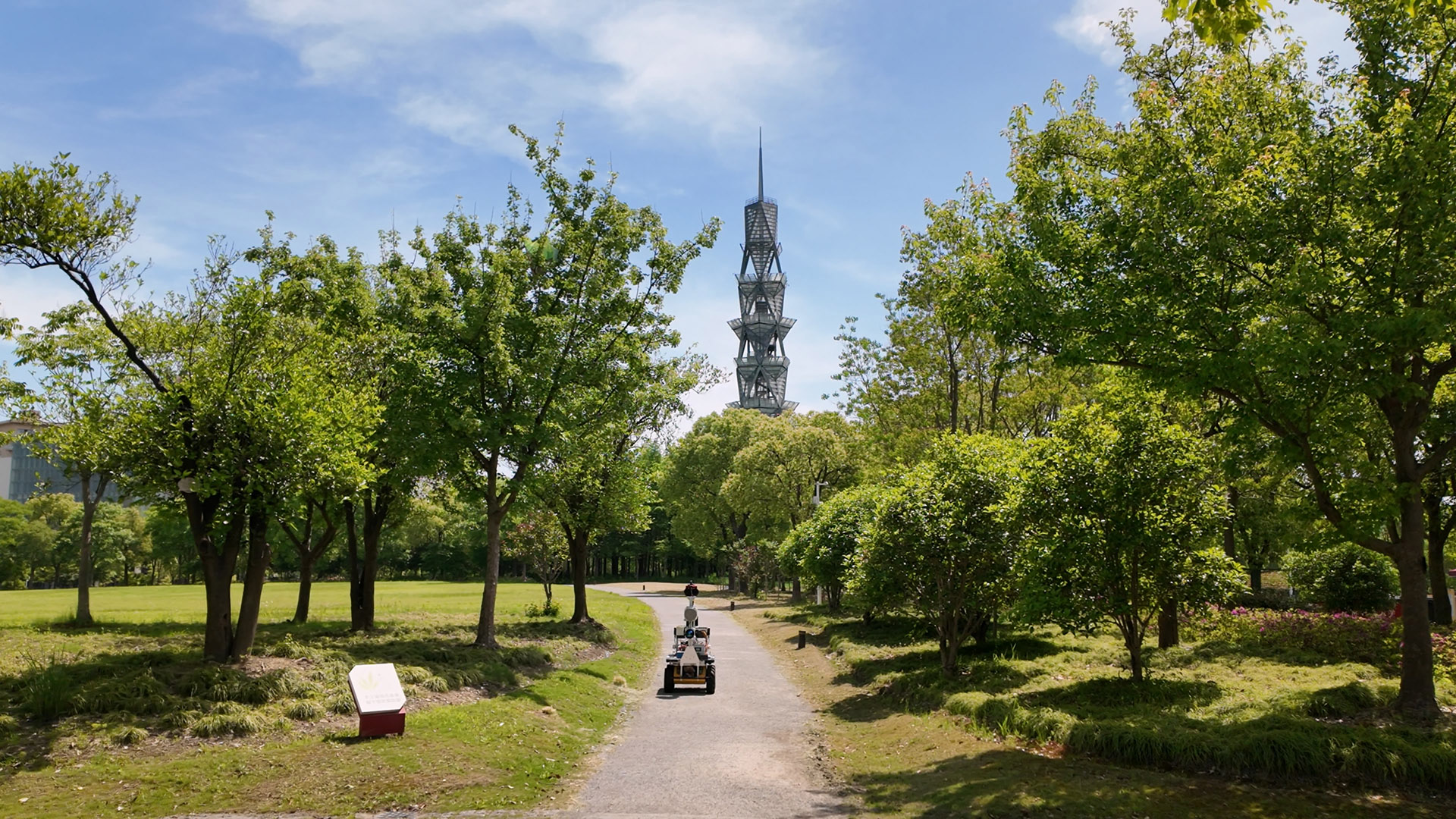}}
	\caption{Mapping robot performing data acquisition on ShanghaiTech campus.}
	\label{fig:outdoor_data_collection_view}
\end{figure}

The experiment was conducted on a quite overcast day, providing even lighting on objects and preventing direct sunlight spots in the collected images. The robot took 26 minutes to traverse the entire route, ultimately collecting 2,618 GiB of data, as shown in Fig. \ref{fig:outdoor_data_collection_gps_result}. A detailed list of the recorded ROS topics is provided in Appendix \ref{appendix:dataset_topics}.

\subsection{LiDAR-based SLAM}
\begin{figure}[htbp]
\centerline{\includegraphics[width=\linewidth]{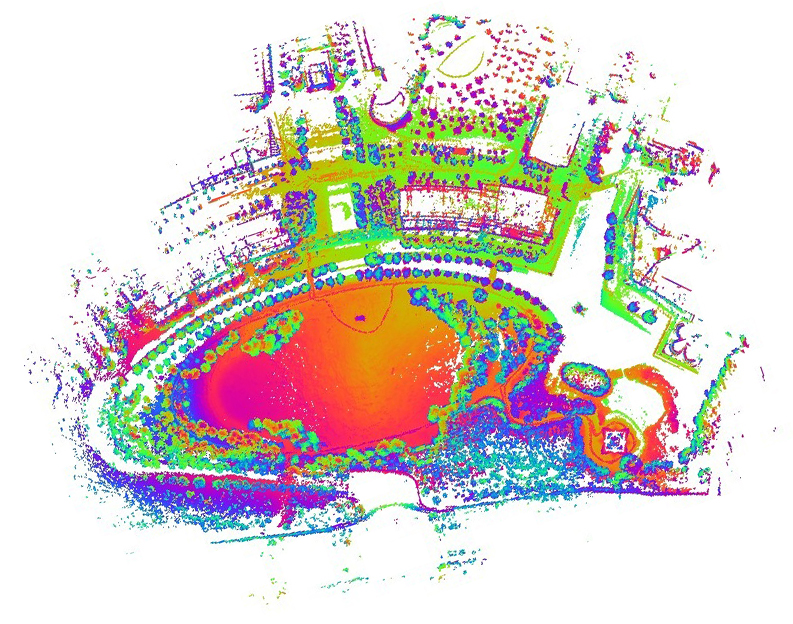}}
\caption{LiDAR Point Cloud Map Generated Using LIO-SAM.}
\label{fig:liosam}
\end{figure}

Our LiDAR SLAM implementation integrates two state-of-the-art algorithms: LIO-SAM \cite{shan2020lio} and MA-LIO \cite{jung2023asynchronous}. For LIO-SAM, we activate loop closure detection and utilize the front-mounted RoboSense Ruby LiDAR sensor. Fig. \ref{fig:liosam} provides an example of the LiDAR point cloud map generated using LIO-SAM. In contrast, the MA-LIO configuration operates without loop closure, instead leveraging data from multiple LiDAR units, including RoboSense and Hesai sensors. 

\begin{figure}[htbp]
	\centering
	\includegraphics[width=0.32\linewidth]{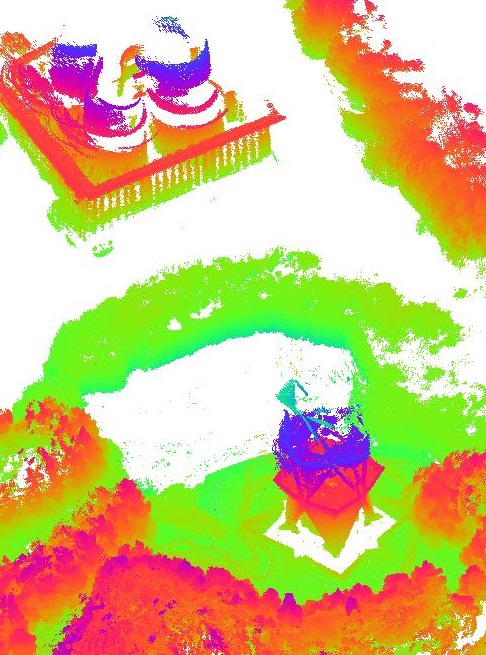}
	\includegraphics[width=0.32\linewidth]{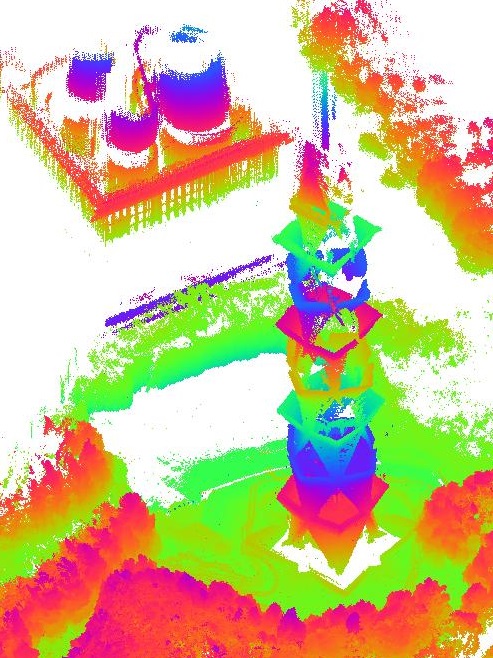}
	\includegraphics[width=0.32\linewidth]{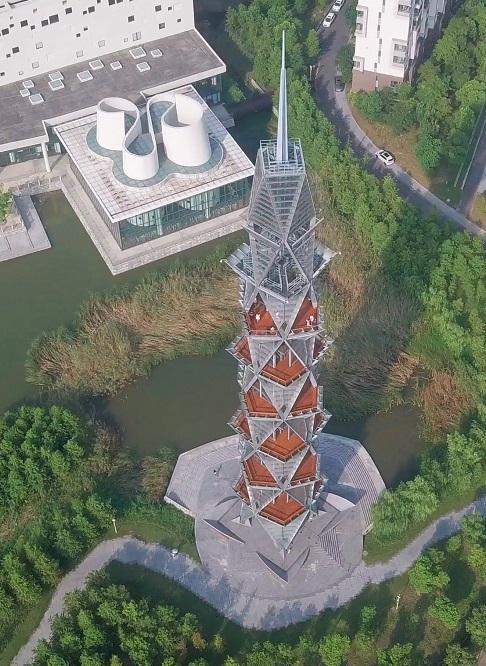}
	\caption{Mapping the ShanghaiTech Tower: MA-LIO without vertical LiDAR (left), MA-LIO with vertical LiDAR (middle), and photo (right).}
	\label{fig:tower}
\end{figure}

As demonstrated in Fig. \ref{fig:tower}, the multi-sensor fusion with the vertical Robosense Ruby LiDAR  allows the system to accurately capture the facades of high-rise buildings and the sides of towers. The expanded sensor array enabled the reconstruction of a more complete 3D map, which proves especially beneficial in urban areas with prominent vertical structures.

\subsection{Visual SLAM}
For visual SLAM, we employ ORB-SLAM3 \cite{ORB3} in both monocular-inertial and stereo-inertial configurations. Our setup utilizes the front-left camera for monocular mode and the front stereo camera pair for stereo mode. We conduct a comparative analysis of the results from these two configurations to evaluate their relative performance. 

\begin{figure}[htbp]
\centerline{\includegraphics[width=\linewidth]{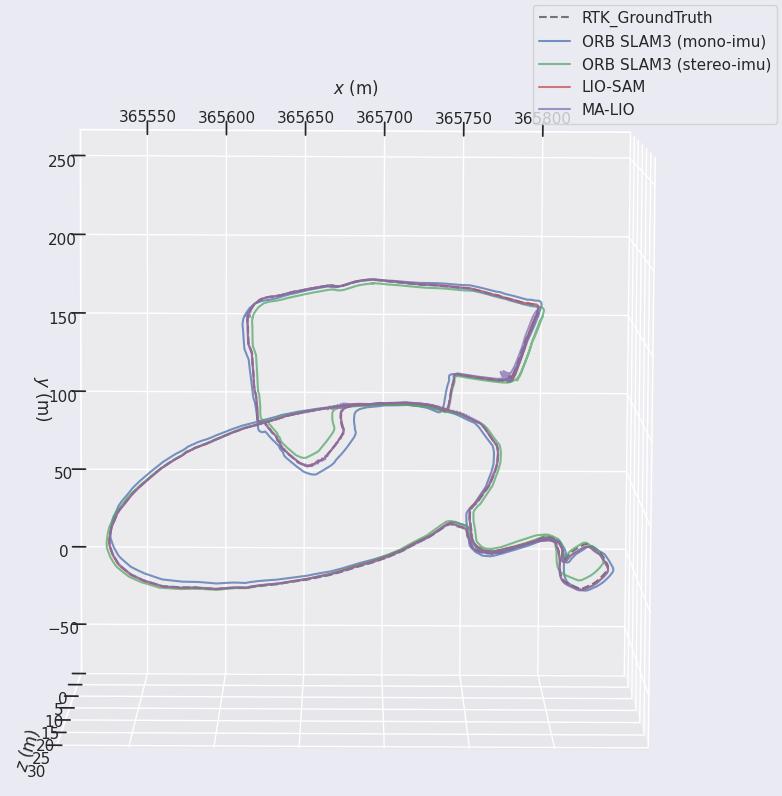}}
\caption{Trajectory comparison generated by LIO-SAM, MA-LIO, and ORB-SLAM3, visualized using evo, alongside the ground truth trajectory from RTK.}
\label{fig:evo}
\end{figure}

\begin{figure*}[htbp]
\centerline{\includegraphics[width=.9\linewidth]{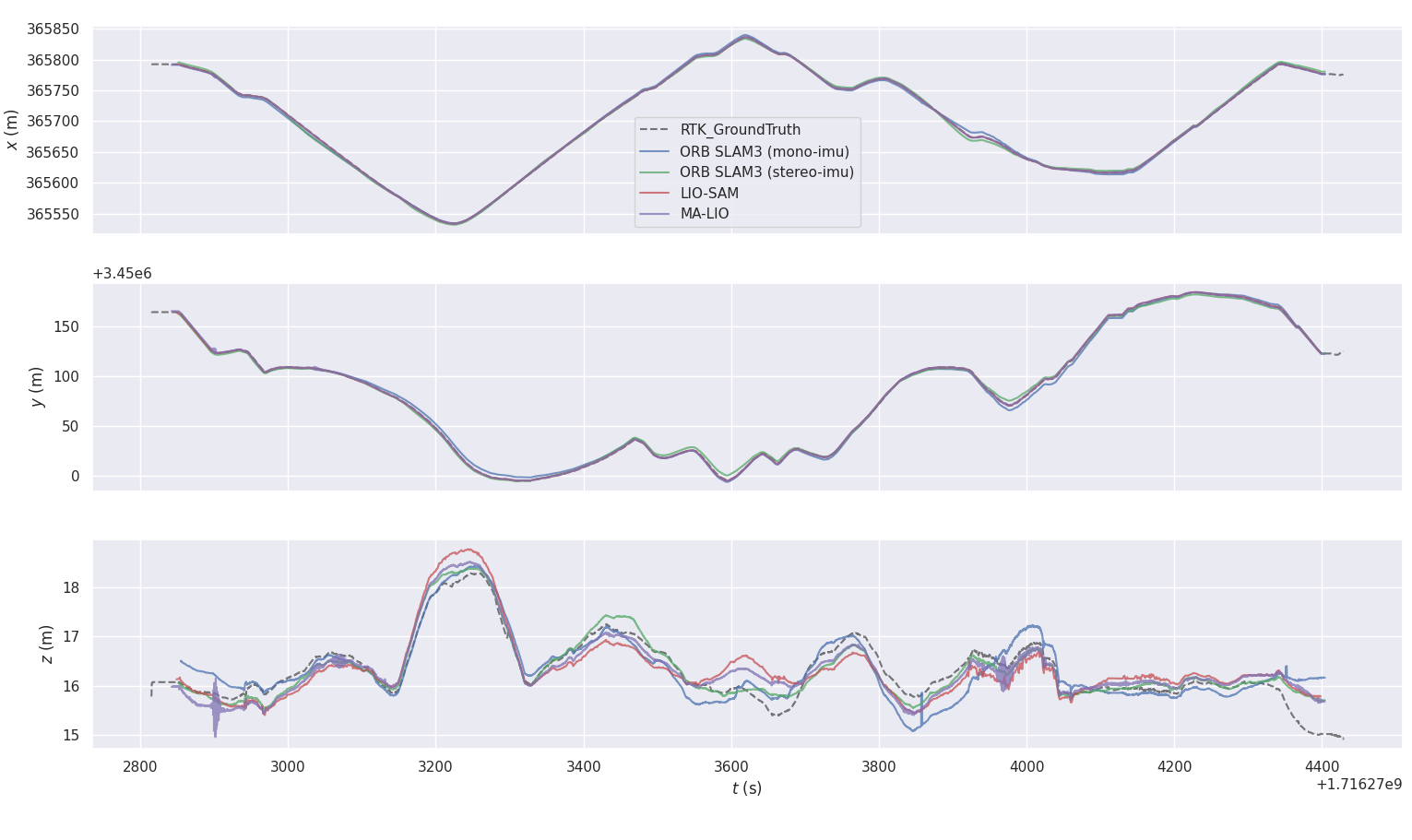}}
\caption{Outdoor dataset trajectory in the x, y, and z coordinates.}
\label{fig:evoxyz}
\end{figure*}

We use evo \cite{grupp2017evo} to visualize and compare the pose estimation results from different SLAM algorithms. For visualization, we apply evo's alignment function to adjust both the translation and scale of the estimated trajectories relative to the ground truth from the RTK system. Fig. \ref{fig:evo} shows the aligned trajectory produced by LIO-SAM, MA-LIO, and ORB-SLAM3, overlaid with the ground truth trajectory from the RTK system. Fig. \ref{fig:evoxyz} provides a detailed breakdown of the trajectory in the x, y, and z dimensions, offering deeper insight into the variations in pose estimation over time across the different SLAM systems. Table \ref{table:slam} compares the absolute pose error (APE) and relative pose error (RPE) of LIO-SAM, MA-LIO, and two configurations of ORB-SLAM3, underscoring the superior performance of LiDAR-based methods in environments with significant visual challenges.

Despite the assistance of an inertial measurement unit (IMU), visual SLAM systems such as ORB-SLAM3 still suffer from scale drift, particularly in monocular configurations where depth information is lacking. While the stereo-inertial setup mitigates this to some extent by providing direct depth measurements, scale drift remains a challenge, especially in long-duration or visually-degraded scenarios. LiDAR-based SLAM methods, by contrast, exhibit far more stable performance in such environments due to the precise distance measurements inherent to LiDAR sensors.

\begin{table}[htbp]
\caption{Comparison of absolute pose error (APE) and relative pose error (RPE) across different SLAM algorithms.}
\centering
\resizebox{\linewidth}{!}{
\begin{tabular}{c|c|c|c|c}
\hline
       & LIO-SAM & MA-LIO & ORB3 mono imu & ORB3 stereo imu\\ \hline
APE (m) & 0.789   & 0.935  & 4.611 & 22.055 \\
RPE (m) & 1.307   & 0.114  & 0.036 & 0.041\\ \hline
\end{tabular}
}
\label{table:slam}
\end{table}


\section{Future Work}
\label{sec:futurework}

This paper introduced the ShanghaiTech Mapping Robot and the outdoor dataset collected with it. The future work ahead of us can be split into two categories: further improving the robot and data collected as well as using the datasets for various exciting research areas.

We are currently working on upgrading the sensor setup of the robot by adding more event cameras (to stereo front and monocular left, right, back, up and down). We will also improve the data quality by filtering out the interfered points identified in Section \ref{sec:Interference}. 

Then we will collect the definitive ShanghaiTech Robotic Dataset, with up to 2 hours indoor and outdoor sequences, augmented by data from other robot modalities (e.g. drones and quadrupeds). This dataset will also feature ground truth 3D point cloud data collected with a Faro 3D scanner. Data will also be collected at different times of the day, including during lunch rush-hour, with lots of students as dynamic objects. We also plan to collect datasets in other environments, such as industry, ancient architecture or agriculture settings.

In order to cover bigger outdoor areas we have been working on mounting the ShanghaiTech Mapping Robot on an electric flatbed car, increasing the speed of the data acquisition platform from 1 m/s to roughly 7 m/s \cite{flatbed}. Data from this setup will be included in the ShanghaiTech Robotic Dataset as well.

We may also pursue making the ShanghaiTech Mapping Robot fully autonomous in the future, to enable the capture of datasets along exactly the same trajectory, which might be useful for certain types of approaches \cite{8968283}, such as lifelong SLAM \cite{shi2020we}. 

The datasets collected with our system will enable a wide range of research topics. The foremost application and one of the main motivations for building this robot is to evaluate SLAM algorithms in great depth. For this we have developed the SLAM Hive Benchmarking Suite~\cite{liu2024benchmarkingslamalgorithmscloud, yang2023slam}, a docker-based distributed system to run, evaluate and analyze SLAM systems with various datasets, based on their ground truth paths. The datasets collected with our robot open up the full potential of SLAM Hive, by providing high-resolution, high-frame rate data from various sensors in various configurations, thus, for the first time, enabling a thorough, automated comparison of SLAM algorithms of various flavours (e.g. LiDAR-based, RGB, RGB-D, event, radar, ...) with various sensor configurations (e.g. front- vs. side-facing sensors), sensor combinations and algorithm configurations (e.g. number of particles, number of features, frame rate, resolution). We hope that this will give important insights into SLAM algorithms and provide a powerful tool for the design and evaluation of future SLAM approaches. 

Down the line we furthermore plan to fully label the ground truth FARO point cloud. Together with the ground truth path of the robot we can then project those labels back to all sensor data from the robot. We hope that this can further drive the research in areas such as segmentation or object detection and tracking. The datasets can then also be used in an extended version of SLAM Hive to benchmark such kind of algorithms. 

In general, given the data-richness of our dataset, we hope that researchers will use it to develop and test all kinds of interesting algorithms, especially w.r.t to the fusion of unconventional sensors, e.g. event with ultrasound and IMU data, or event with radar, etc. 

Finally, we will also utilize the datasets to generate osmAG maps (open street map Area Graph) \cite{feng2023osmag}, which we have utilized for Robust Lifelong Indoor Localization \cite{xie2023robust} and to enable LLMs (Large Language Models) to interpret maps \cite{xie2024empowering}.


\section{Conclusions}
\label{sec:conclusions}

In this paper we have presented the ShanghaiTech Mapping Robot and the outdoor dataset collected with it. The robot features a never-seen before richness of high-quality sensors, including RGB and RGB-D cameras, IR and event cameras, LiDARs, IMUs, ultrasound and mmWave range sensors as well as an GNSS RTK receiver. A 16-node cluster was constructed to enable the collection of all of this data. The mechanical structure is very rigid to enable the precise extrinsic calibration of the sensors. A hardware-synchronization approach ensure the proper triggering and time-stamping of most of the sensor data. We have described the procedures of dataset collection and our post-processing steps. Our investigation of the interference of the active IR-based sensors showed that just two of the front LiDAR suffer from a limited amount of range readings with added noise. 

\begin{table*}[htbp]
\centering
\caption{Sensor Data Statistics of Outdoor Dataset}
\label{table:dataset_topics_detail}
\resizebox{\textwidth}{!}{%
\begin{tabular}{l|l|l|c|c|c|c|l}
\hline
\textbf{Sensor Type} & \textbf{Model} & \textbf{Topic} & \textbf{Topic Count} & \textbf{Messages} & \textbf{Data Rate} & \textbf{Duration} & \textbf{Message Type} \\
\hline
RGB Camera & FLIR Grasshopper3 & /GS3\_\{\{front,back,right,left,top\}\_\{L,R\},down\}/image\_raw & 11 & 46,860 & 30 Hz & 26:01 & sensor\_msgs/Image \\
& & /GS3\_\{\{front,back,right,left,top\}\_\{L,R\},down\}/camera\_info & 11 & 46,860 & 30 Hz & 26:01 & sensor\_msgs/CameraInfo \\
& FLIR LadyBug5+ & /ladybug/camera\{0-5\}/bayer\{0-3\}/compressed & 24 & 23,430 & 15 Hz & 26:01 & sensor\_msgs/CompressedImage \\
\hline
RGB-D Camera & Azure Kinect DK & /Kinect\_\{front,back,right,left,top\}/rgb/image\_raw/compressed & 5 & 7,989-7,992 & 5 Hz & 26:38 & sensor\_msgs/CompressedImage \\
& & /Kinect\_\{front,back,right,left,top\}/rgb/camera\_info & 5 & 7,989-7,992 & 5 Hz & 26:38 & sensor\_msgs/CameraInfo \\
& & /Kinect\_\{front,back,right,left,top\}/depth/image\_raw & 5 & 7,989-7,992 & 5 Hz & 26:38 & sensor\_msgs/Image \\
& & /Kinect\_\{front,back,right,left,top\}/depth/camera\_info & 5 & 7,989-7,992 & 5 Hz & 26:38 & sensor\_msgs/CameraInfo \\
& & /Kinect\_\{front,back,right,left,top\}/imu & 5 & 2,651,072 (typical) & 1600 Hz (typ.) & 26:38 & sensor\_msgs/Imu \\
\hline
IR Camera & FLIR A315 & /A315\_\{L,R\}/image\_raw & 2 & 96,369-96,373 & 60 Hz & 26:46 & sensor\_msgs/Image \\
\hline
Event-based Camera & EvC3A & /EvC3A\_front\_\{L,R\}/events & 2 & 133,007-133,095 & N/A & 26:41 & event\_camera\_msgs/EventPacket \\
\hline
Spinning LiDAR & Robosense Ruby & /ruby\_\{front,back\}/rslidar\_points & 2 & 15,966-15,969 & 10 Hz & 26:36 & sensor\_msgs/PointCloud2 \\
& & /ruby\_\{front,back\}/rslidar\_packets & 2 & 19,163,367 (typ.) & N/A & 26:36 & rslidar\_msg/RslidarPacket \\
& Robosense Bpearl & /bpearl\_\{front,back\}/rslidar\_points & 2 & 15,963-15,965 & 10 Hz & 26:36 & sensor\_msgs/PointCloud2 \\
& & /bpearl\_\{front,back\}/rslidar\_packets & 2 & 4,807,801 (typ.) & N/A & 26:36 & rslidar\_msg/RslidarPacket \\
& Hesai PandarQT & /hesai/pandar & 1 & 16,031 & 10 Hz & 26:41 & sensor\_msgs/PointCloud2 \\
\hline
Solid-state LiDAR & Livox Avia & /livox/lidar & 1 & 16,038 & 10 Hz & 26:43 & sensor\_msgs/PointCloud2 \\
& & /livox/imu & 1 & 325,608 & 200 Hz & 26:43 & sensor\_msgs/Imu \\
& Neuvition M1 & /neuvition\_cloud & 1 & 1,489 & 10 Hz & 2:27 & sensor\_msgs/PointCloud2 \\
& & /neuvition\_image/compressed & 1 & 859 & 5 Hz & 2:27 & sensor\_msgs/CompressedImage \\
\hline
IMU & Xsens MTi-630R & /mti/imu/data & 1 & 624,797 & 400 Hz & 26:02 & sensor\_msgs/Imu \\
& & /mti/filter/free\_acceleration & 1 & 624,797 & 400 Hz & 26:02 & geometry\_msgs/Vector3Stamped \\
& & /mti/filter/quaternion & 1 & 624,797 & 400 Hz & 26:02 & geometry\_msgs/QuaternionStamped \\
& & /mti/imu/acceleration & 1 & 624,797 & 400 Hz & 26:02 & geometry\_msgs/Vector3Stamped \\
& & /mti/imu/angular\_velocity & 1 & 624,797 & 400 Hz & 26:02 & geometry\_msgs/Vector3Stamped \\
& & /mti/imu/dq & 1 & 624,797 & 400 Hz & 26:02 & geometry\_msgs/QuaternionStamped \\
& & /mti/imu/dv & 1 & 624,797 & 400 Hz & 26:02 & geometry\_msgs/Vector3Stamped \\
& & /mti/imu/mag & 1 & 624,797 & 400 Hz & 26:02 & geometry\_msgs/Vector3Stamped \\
& & /mti/imu/time\_ref & 1 & 624,797 & 400 Hz & 26:02 & sensor\_msgs/TimeReference \\
& & /mti/pressure & 1 & 624,797 & 400 Hz & 26:02 & sensor\_msgs/FluidPressure \\
& & /mti/temperature & 1 & 624,797 & 400 Hz & 26:02 & sensor\_msgs/Temperature \\
\hline
Ultrasonic Sensor & KS103 & /ultrasound/range/\{1-13\} & 13 & 1,387 & 1 Hz & 26:47 & sensor\_msgs/Range \\
\hline
mmWave Radar & TI IWR6843AOPEVM & /mmWave\_\{front,back,left,right,top\}/radar\_scan\_pcl & 5 & 48,066-48,074 & 30 Hz & 26:42 & sensor\_msgs/PointCloud2 \\
& & /mmWave\_\{front,back,left,right,top\}/radar\_scan & 5 & 209,123 (typ.) & N/A & 26:42 & ti\_mmwave\_rospkg/RadarScan \\
\hline
GNSS Receiver & Bynav A1 & /gps/fix & 1 & 8,045 & 5 Hz & 26:49 & sensor\_msgs/NavSatFix \\
& & /gps/bestpos & 1 & 8,045 & 5 Hz & 26:49 & novatel\_gps\_msgs/NovatelPosition \\
& & /gps/bestvel & 1 & 8,045 & 5 Hz & 26:49 & novatel\_gps\_msgs/NovatelVelocity \\
& & /gps/gpgga & 1 & 8,045 & 5 Hz & 26:49 & novatel\_gps\_msgs/Gpgga \\
& & /gps/gprmc & 1 & 8,045 & 5 Hz & 26:49 & novatel\_gps\_msgs/Gprmc \\
& & /gps/gps & 1 & 7,988 & 5 Hz & 26:49 & gps\_common/GPSFix \\
& & /gps/inspvax & 1 & 8,046 & 5 Hz & 26:49 & novatel\_gps\_msgs/Inspvax \\
\hline
Synchronization & Syncboard & /syncboard/line/\{2,14-17\} & 5 & 7,810-624,786 & 5-400 Hz & 26:40 & std\_msgs/Time \\
& & /syncboard/status & 1 & 1,601 & 1 Hz & 26:40 & mars\_syncboard\_srvs/BoardStatus \\
\hline
Robot Base & Clearpath Husky & /joy\_teleop/joy\_selected & 1 & 37,980 & 24 Hz & 26:41 & sensor\_msgs/Joy \\
& & /status & 1 & 1,602 & 1 Hz & 26:41 & husky\_msgs/HuskyStatus \\
\hline
\multicolumn{3}{r|}{\textbf{Total}} & \textbf{137} & \multicolumn{4}{l}{\textbf{\# of bags: 37, size of bags: 2,618 GiB}} \\
\hline
\end{tabular}%
}
\end{table*}

The paper also presents a preliminary dataset acquired with our ShanghaiTech Mapping Robot. To demo some of its features we used it to compare the pose errors of two LiDAR-based (LIO-SAM and MA-LIO) and two vision-based (ORB3 mono imu and ORB3 stereo imu) SLAM approaches, with the LiDAR-based approaches showing better absolute pose errors than ORB3. We also motivated the inclusion of the vertically-mounted Robosense Ruby LiDAR by highlighting how only its data can provide a good point-cloud representation of tall vertical structures like the ShanghaiTech Tower. 

Overall, this system paper showed what very rich datasets this robot will be able to collect in the future, providing a highly useful source of data that can be of high value in research areas such as SLAM evaluation and SLAM, sensor fusion and all kinds of perception approaches for robotics.

\section{Acknowledgments}

This work was supported by Science and Technology Commission of Shanghai Municipality (STCSM), project 22JC1410700 \textquotedblleft Evaluation of real-time localization and mapping algorithms for intelligent robots\textquotedblright .
This work has also been partially funded by the Shanghai Frontiers Science Center of Human-centered Artificial Intelligence. The experiments of this work were supported by the core facility Platform of Computer Science and Communication, SIST, ShanghaiTech University.

\appendices

\section{Details of ROS topics recorded in experiment}
\label{appendix:dataset_topics}

In Table \ref{table:dataset_topics_detail}, we show for each type and model of sensor the message topics they publish to as well as the total number of messages recorded from them in the dataset. The time duration, data rate, and message type are also listed.

\bibliographystyle{IEEEtran}

\bibliography{References}
	
	\begin{IEEEbiography}[{\includegraphics[width=1in,height=1.25in,clip,keepaspectratio]{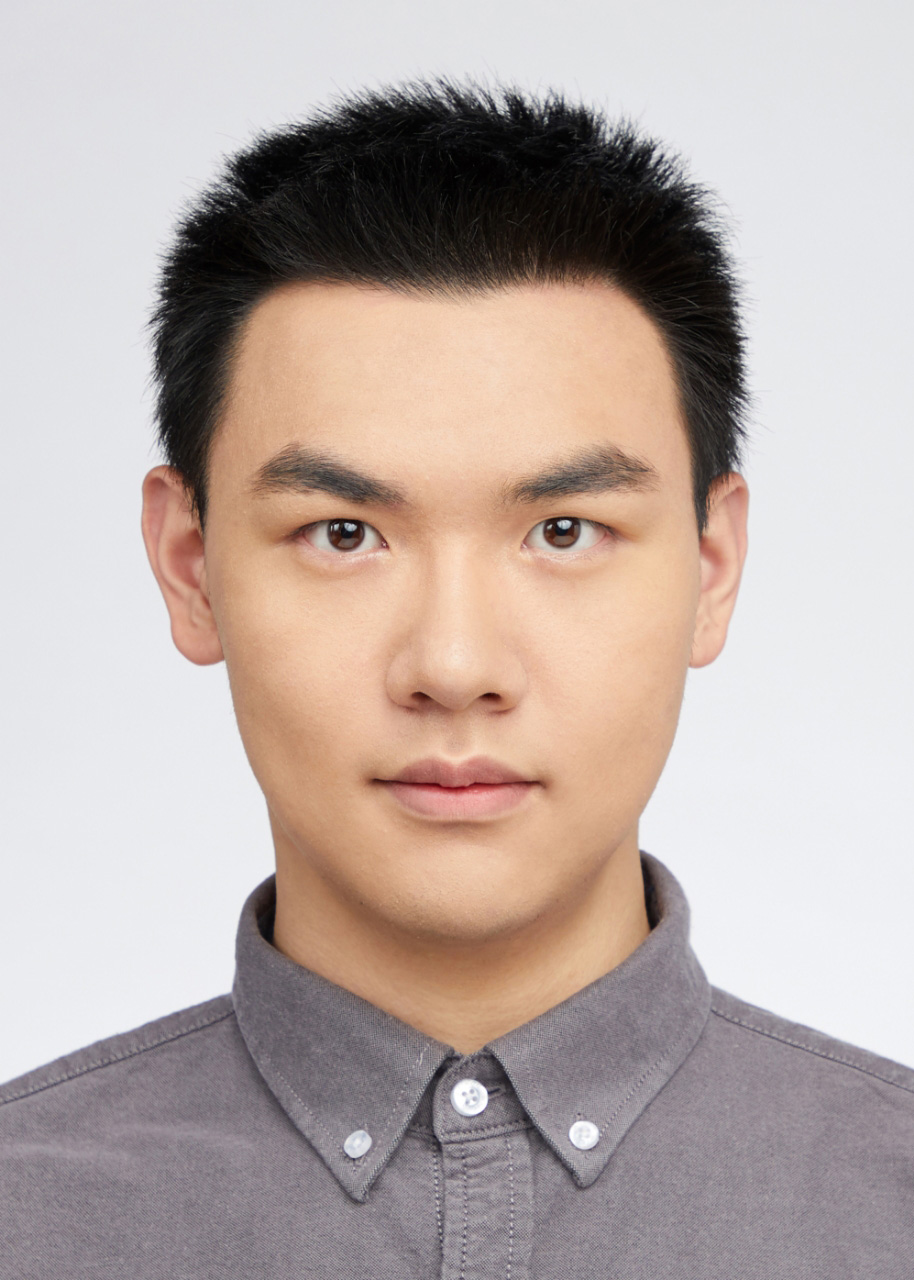}}]{Bowen Xu}
		received his B.S. at ShanghaiTech University in China, and now he is a third-year graduate student at ShanghaiTech University. His research focuses on Robot Mapping and SLAM.
	\end{IEEEbiography}

	\begin{IEEEbiography}[{\includegraphics[width=1in,height=1.25in,clip,keepaspectratio]{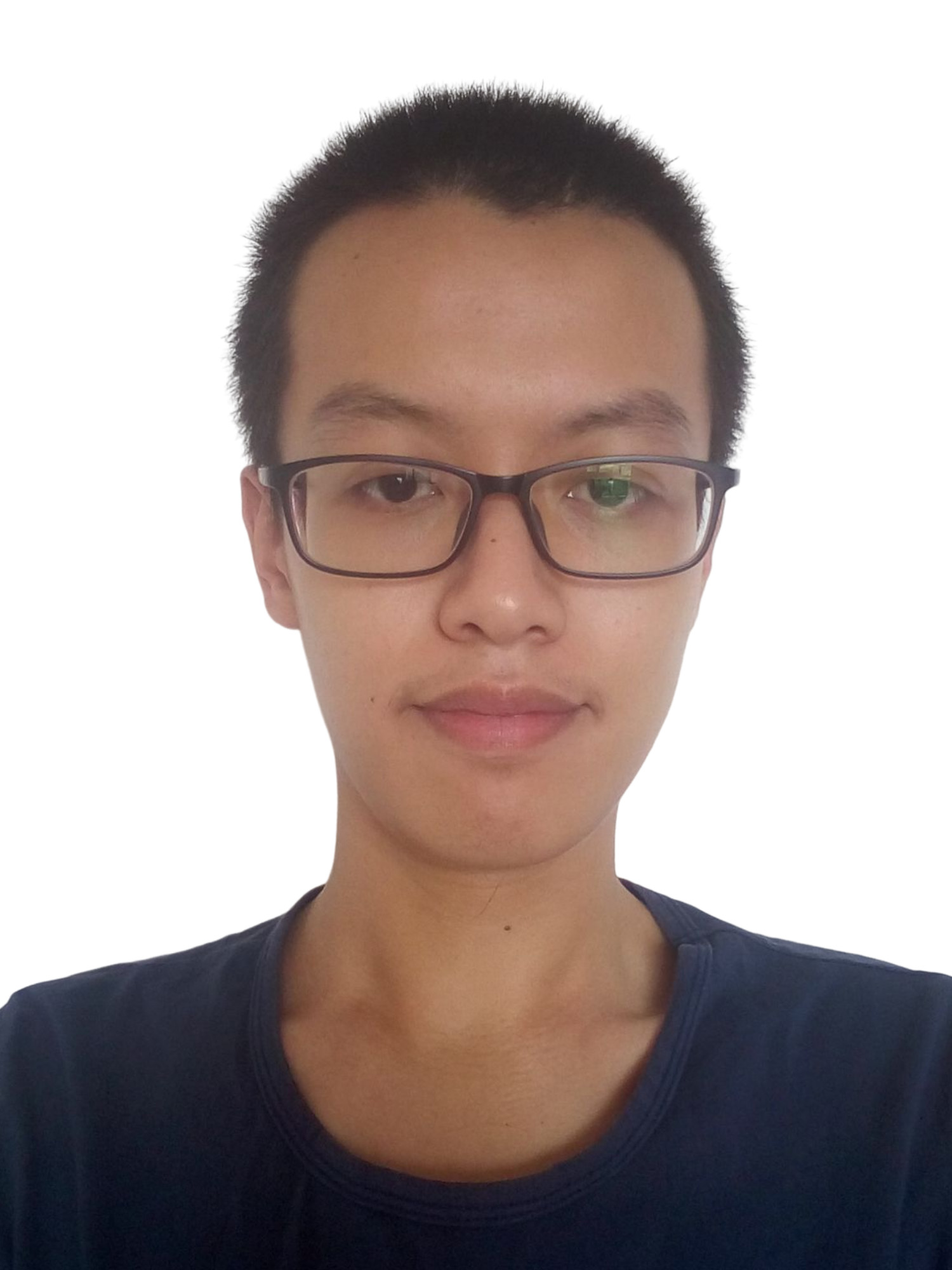}}]{Xiting Zhao}
		  is currently a PhD candidate in Computer Science and Technology at the School of Information Science, ShanghaiTech University, under the guidance of Professor Sören Schwertfeger. Xiting earned his B.S in Computer Science and Technology from the same institution in 2019. His research focuses on SLAM, Robot Mapping, and 3D reconstruction.
	\end{IEEEbiography}

    \begin{IEEEbiography}[{\includegraphics[width=1in,height=1.25in,clip,keepaspectratio]{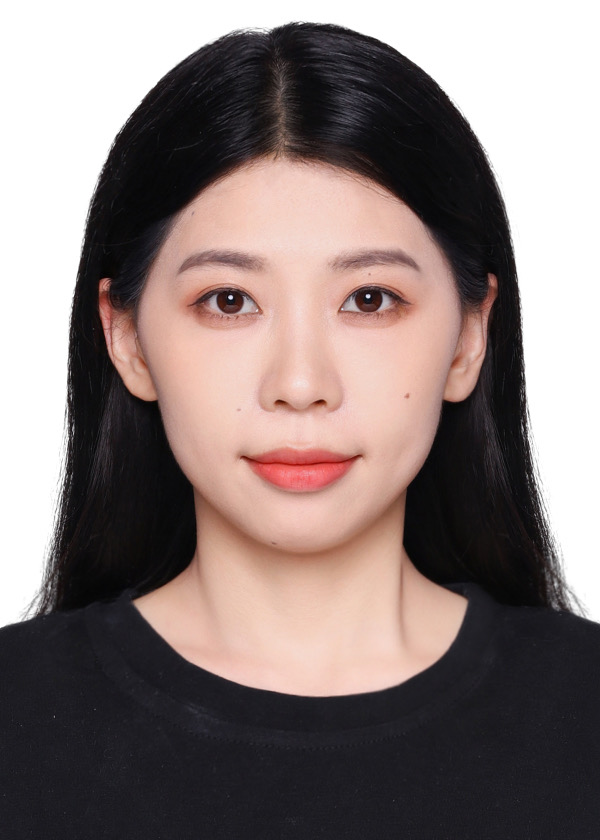}}]{Delin Feng}
		received Master degree in Computer Science and Technology from ShanghaiTech University, Shanghai, China, in 2023. Her research focuses on SLAM, Scene Graph (Map), End2End Autonomous Driving. She is currently working in industry on End2End autonomous driving.
	\end{IEEEbiography}
	
	\begin{IEEEbiography}[{\includegraphics[width=1in,height=1.25in,clip,keepaspectratio]{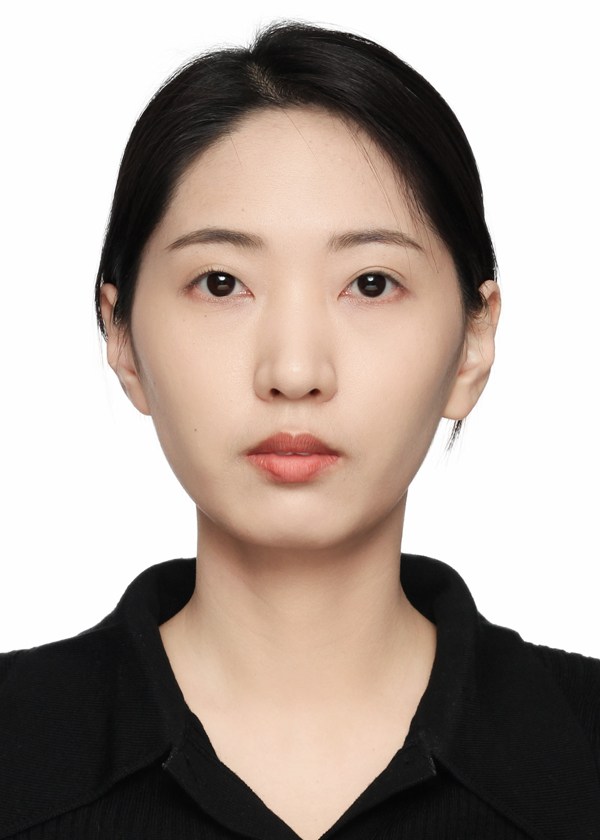}}]{Yuanyuan Yang}
		received Master degree in Computer Science and Technology from ShanghaiTech University, Shanghai, China, in 2023. Her research interests include SLAM evaluation, sensor fusion and mapping for moblie robots. She is currently working in industry on autonomous  driving of intelligent electric vehicles.
	\end{IEEEbiography}
	
	\begin{IEEEbiography}[{\includegraphics[width=1in,height=1.25in,clip,keepaspectratio]{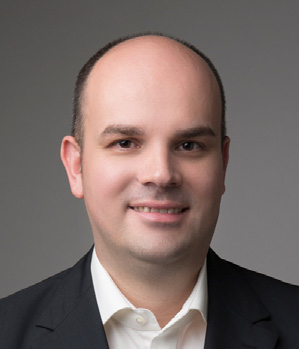}}]{Prof. Dr. S\"oren Schwertfeger}
		 is an Associate Professor
		at ShanghaiTech University, where he joined in 2014. He is the head of the Mobile Autonomous Robotics Systems Lab and a director of the ShanghaiTech Automation and Robotics Center. 
		In 2012 he
		received his Ph.D. in Computer Science from the Jacobs
		University Bremen, researching at the robotics group of Prof.
		Andreas Birk. In 2010
		Dr. Schwertfeger was a guest researcher at the National Institute of Standards
		and Technology (NIST) in Gaithersburg, Maryland, USA. His
		research interest is in robotics, especially intelligent functions for mobile robots.
		Besides his work on mapping, map representation and SLAM, Dr. Schwertfeger is working on mobile
		manipulation, robot autonomy an AI for robotics. He was the general chair of the 2017 IEEE International Symposium
		on Safety, Security, and Rescue Robotics (SSRR), a guest editor of the Journal of Field Robotics and an associate editor of the IEEE Robotics and Automation Magazine (2018-2021). He is also a member of the board of Trustees of the RoboCup Federation.
	\end{IEEEbiography}

\end{document}